# Entropy-guided Retinex anisotropic diffusion algorithm based on partial differential equations (PDE) for illumination correction

U. A. Nnolim

*Department of Electronic Engineering, University of Nigeria Nsukka, Enugu, Nigeria*

### Abstract

*This report describes the experimental results obtained using a proposed variational Retinex algorithm for controlled illumination correction. Two colour restoration and enhancement schemes of the algorithm are presented for drastically improved results. The algorithm modifies the reflectance image using global and local contrast enhancement approaches and gradually removes the residual illumination to yield highly pleasing results. The proposed algorithms are optimized by way of simultaneous perceptual quality metric (PQM) stabilization and entropy maximization for fully automated processing solving the problem of determination of stopping time. The usage of the HSI or HSV colour space ensures a unique solution to the optimization problem unlike in the RGB space where there is none (forcing manual selection of number of iteration. The proposed approach preserves and enhances details in both bright and dark regions of underexposed images in addition to eliminating the colour distortion, over-exposure in bright image regions, halo effect and grey-world violations observed in Retinex-based approaches. Extensive experiments indicate consistent performance as the proposed approach exploits and augments the advantages of PDE-based formulation, performing illumination correction, colour enhancement correction and restoration, contrast enhancement and noise suppression. Comparisons shows that the proposed approach surpasses most of the other conventional algorithms found in the literature.*

## 1. Introduction

Image enhancement area of image processing is replete with numerous algorithms which range from simple to complex formulations. Earlier linear, statistics-based contrast enhancement algorithms were initially developed for greyscale image contrast enhancement [1]. However, relatively recent algorithms utilize advanced and complex schemes for enhancement of colour images.

The successful rendering of high dynamic range (HDR) images on low dynamic range display (LDR) devices is an ongoing and active field of study. The algorithms for performing this task are referred to as tonal mapping operators (TMOs) and the most popular include the Homomorphic filter [2], and Retinex [3] algorithms. Numerous variants of the latter exceed the former due to the multiple processes incorporated into the formulation. However, the Retinex still has its drawbacks such as halos, greyish tint for certain images and colour distortion and fading. Later works attempted solutions using colour restoration functions and operating in alternative perceptual colour spaces [4], which yielded dramatic colour results. Also, Quaternion Fourier Transform methods have been utilized for colour enhancement [5] [6], though with mixed results. Other alternative approaches include usage of Particle Swarm Optimization (PSO), Genetic Algorithm (GA), Wavelets, Dynamic Stochastic Resonance (DSR) [7].Additionally, the problems of over-exposure of bright image regions and visual halos still persist. Thus, several formulations in the literature have attempted to solve this problem in various ways with varying degrees of success.

It is impossible for an image processing algorithm to work best for all known images though reasonable results for most images is expected. This is experienced with algorithms such as the contrast limited adaptive histogram equalization (CLAHE) [8], Retinex and Homomorphic filters. However, these algorithms are closed-form solutions, difficult to control their effects and thus, cannot adapt to subtle requirements as required in HDR imaging, which are evaluated using the human visual system (HVS).

Partial differential equation (PDE)-based image processing has matured with established works such as Perona and Malik's Anisotropic Diffusion (AD) [9], Rudin, Osher and Fatemi's Total Variation Regularization [10] and Shock filter [11] which were initially designed for filtering Additive White Gaussian Noise (AWGN). However, they have also been applied (especially AD) to other areas in image processing [12] [13] [14] [15].

The utilization of PDE-based image enhancement formulation is not new as seen in [16]. However, in this work, we present results obtained by combining various useful algorithms with a modified process to dramatically improve results not possible to obtain using any of the individual algorithms. Additionally, due to the difficulty of determining the stopping time of the algorithm in RGB space, we resolve this issue by solving for the illumination using only the intensity channel. Furthermore, reliable no-reference image metrics are utilized in optimizing the algorithm for best visual and quantitative results.

The motivation for this work is the development of a fully automated algorithm that would yield consistent results by preserving or enhancing colour and details in both bright and dark image regions respectively. The PDE-based framework enables the gradual processing of images by regulating the contributions of multiple processes within the framework via weighting parameters.

## 2. PDE formulation for image enhancement

The PDE-based formulation proposed in [16] [17] forms the basis for this work. Normally, two combined processes (in this case, smoothing, $F_s()$ and enhancement $F_e()$ functions) acting on a continuous initial image, $I(x, y, t)$ to yield the process shown in (1) where $\lambda$ as a control parameter that regulates the amount of smoothing (using the AD term) with respect to enhancement.

$$\frac{\partial I(x,y,t)}{\partial t} = \lambda F_s\big(I(x,y,t)\big) + F_e\big(I(x,y,t)\big) \qquad (1)$$

$$\frac{\partial I(x,y,t)}{\partial t} = \lambda \|\nabla I(x,y,t)\| \text{div}\left(\frac{\nabla I(x,y,t)}{\|\nabla I(x,y,t)\|}\right) + f\big(I(x,y,t)\big) - I(x,y,t) \qquad (2)$$

The function, $f\big(I(x,y,t)\big)$ in (2) can be any contrast enhancement function though in the original formulation proposed by [16] it is a histogram modification or equalization (HE) transformation function. Other functions, both simple and complex, are employed by other authors to achieve contrast enhancement [18] [19] [20] [21] [6]-[9].

This model is relatively straight-forward and may work with certain images but will yield distorted results for images with uneven illumination. Thus, we modify for such images to improve results.

## 3. Proposed PDE model

Using the base illumination-reflectance model [1], given as;

$$I(x, y) = L(x, y).R(x, y) \qquad (3)$$

Using a logarithm operation transforms the multiplicative relationship between the illumination component, $L(x, y)$ and the reflectance component, $R(x, y)$ into an additive one

$$\log[I(x, y)] = \log[L(x, y)] + \log[R(x, y)] \qquad (4)$$

Leading to;

$$i = l + r \qquad (5)$$

Reformulating into PDE form, we obtain the expression

$$\frac{\partial i(x,y,t)}{\partial t} = \alpha\big(f\{r(x, y, t)\} - i(x, y, t)\big) + \frac{\beta(i(x,y,t) - \mu)}{\sigma} + \lambda g\big(\nabla i(x, y, t)\big)\nabla i(x, y, t) \qquad (6)$$

Where the control parameters; $\alpha$, $\beta$ and $\lambda$ control the illumination correction ($f\big(I(x, y, t)\big)$ here is a local-global-enhancement algorithm), colour correction and smoothing (AD) terms respectively. The term, $\alpha$ is related to the number of iterations in the form;

$$\alpha \propto \frac{1}{N_{iterations}} \qquad (7)$$

Additionally, we test this relationship by processing a typical image with the algorithm 100 iterations at varying values for $\alpha$. The results are shown in Fig. 1.

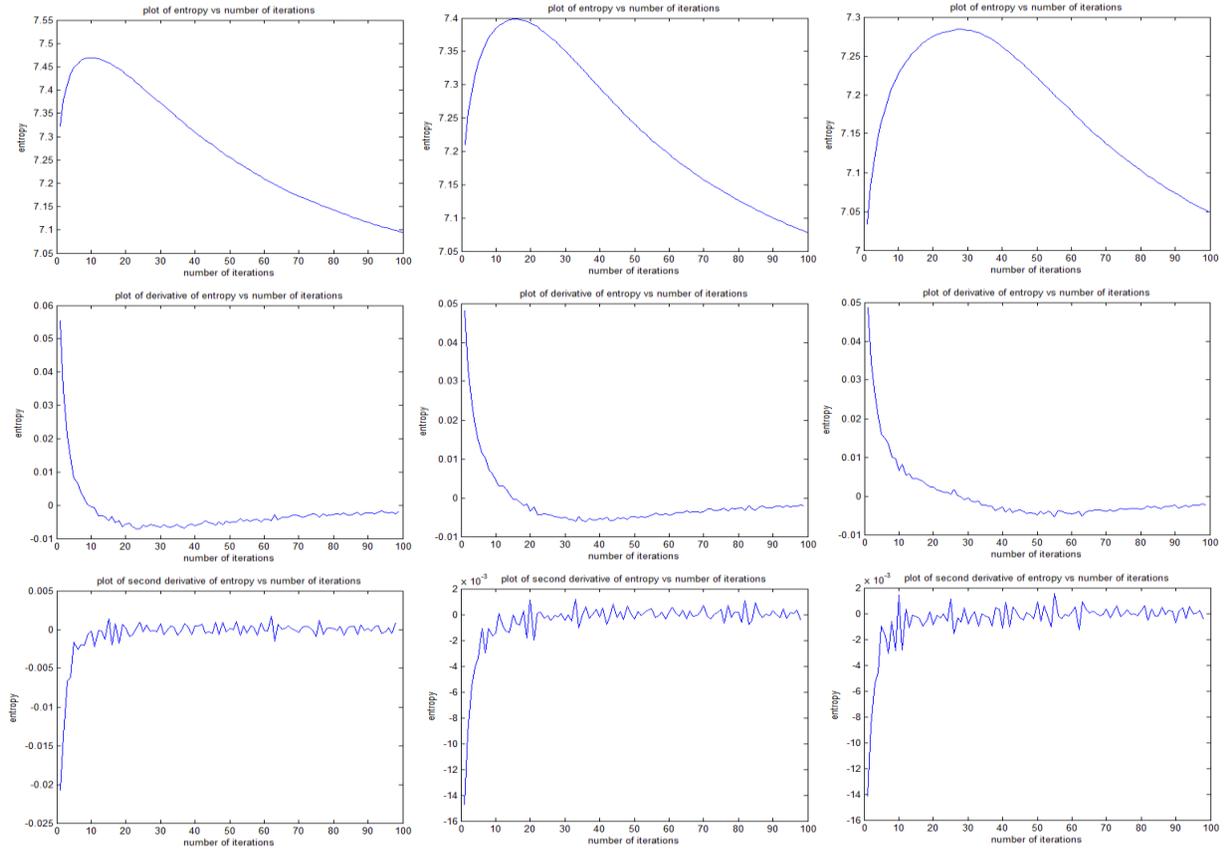

(a)

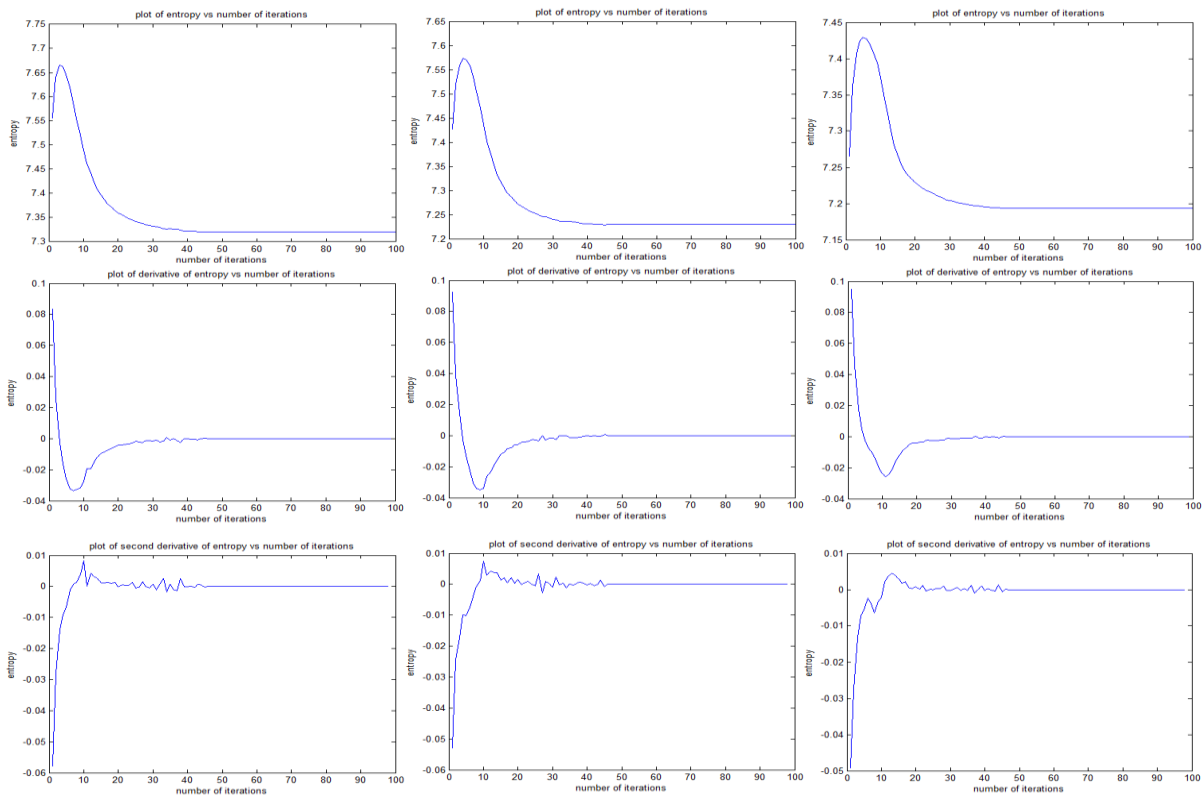

(b)

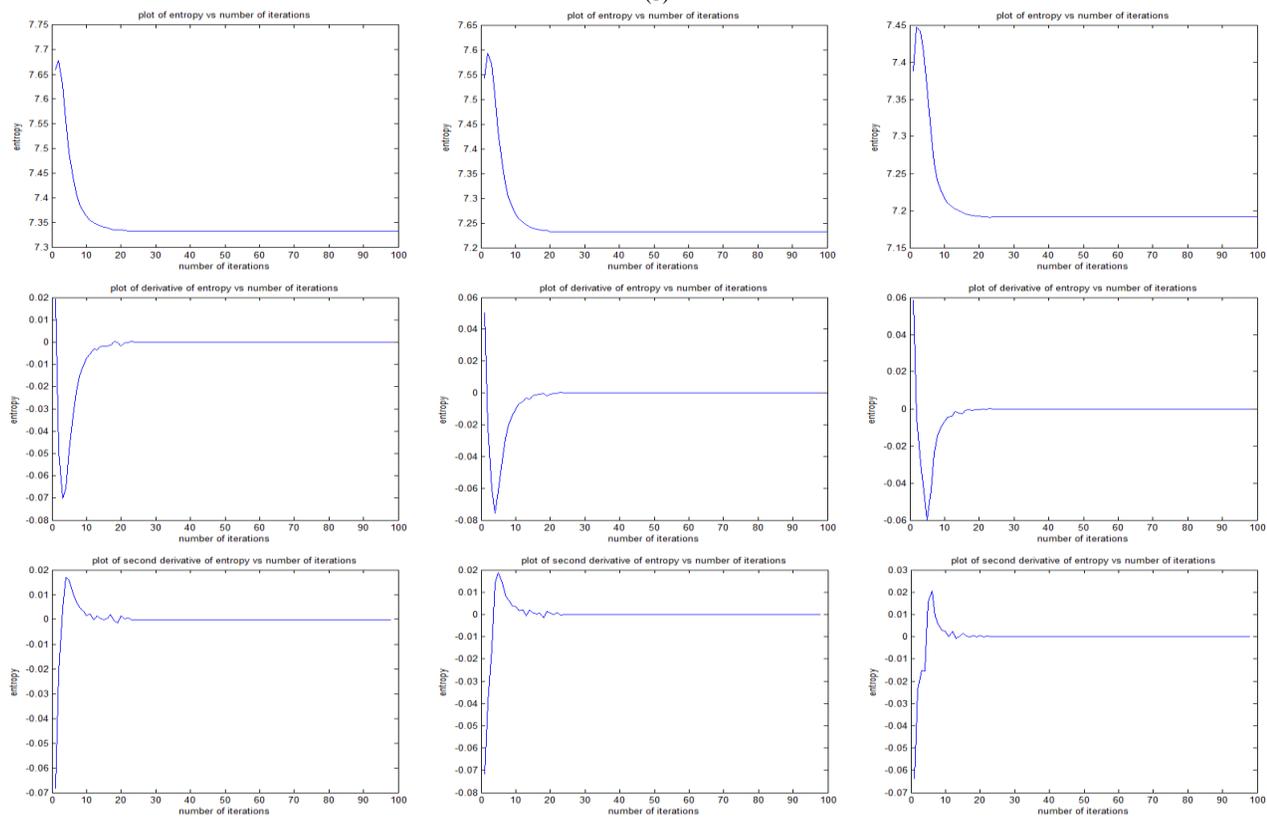

(c)

**Fig. 1** Entropy, first derivative and second derivative of entropy with respect to number of iterations for R, G and B channels for Cathedral image processed with proposed algorithm2 (PA-2) PDE_MSR_CLAHE with RGB-IV colour enhancement algorithm using α = 0.1, 0.5 and 1 for 100 iterations

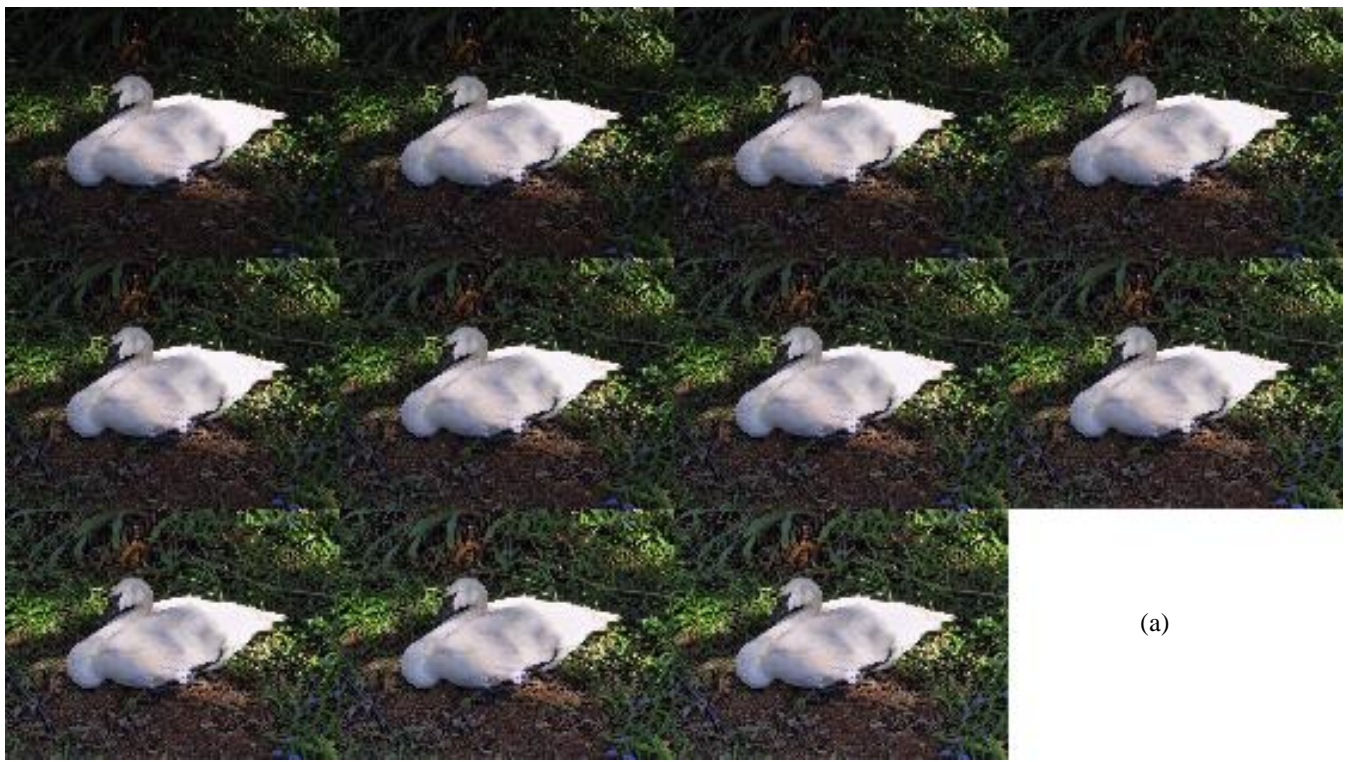

(a)

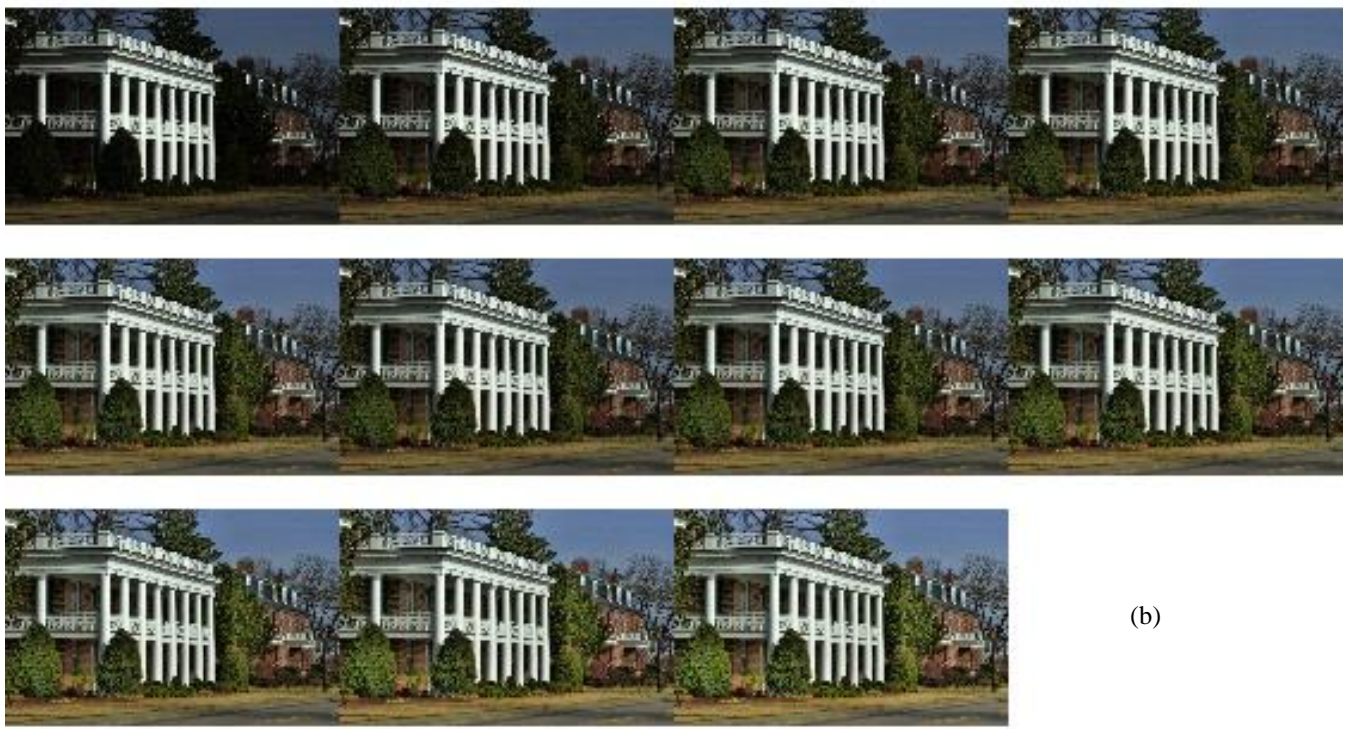

(b)

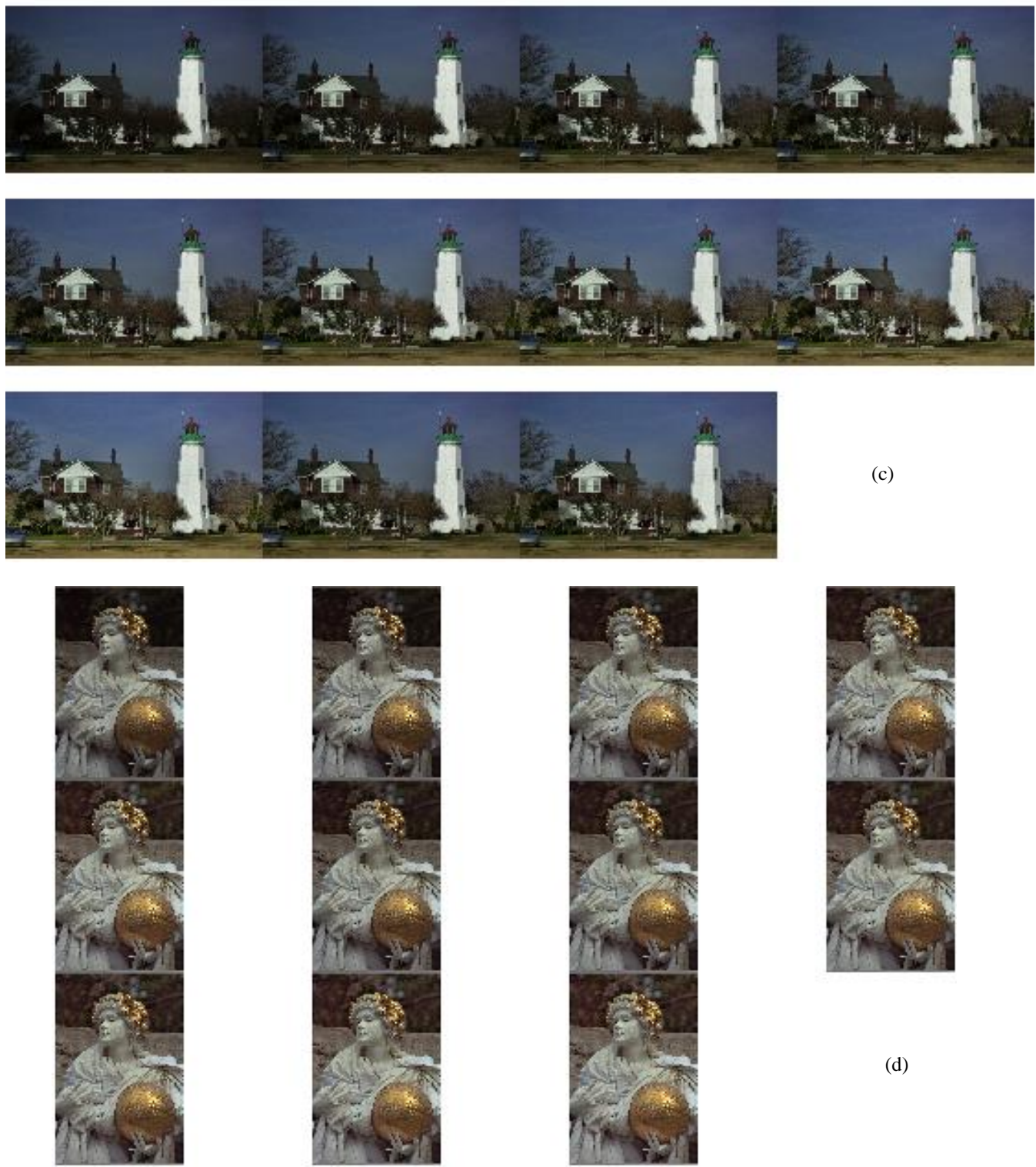

(c)

(d)

**Fig. 2** (a) Swan (b) White house (c) House tower (d) Statue image results using PA for α values ranging from 0.1 to 1, compared with the original image

## 4.    Additional experiments and comparisons

The main quantitative measures used include image entropy, Perceptual Quality Metric (PQM) [22], Colourfulness (C) [23], Colour Enhancement Measurement (EMEC) [5], Contrast Enhancement Factor (CEF/F) [22], Average Gradient (AG) and Hue Deviation Index (HDI) [24]. These are tested to obtain a reliable metric consistent with visual results. The results are shown in Fig. 3.

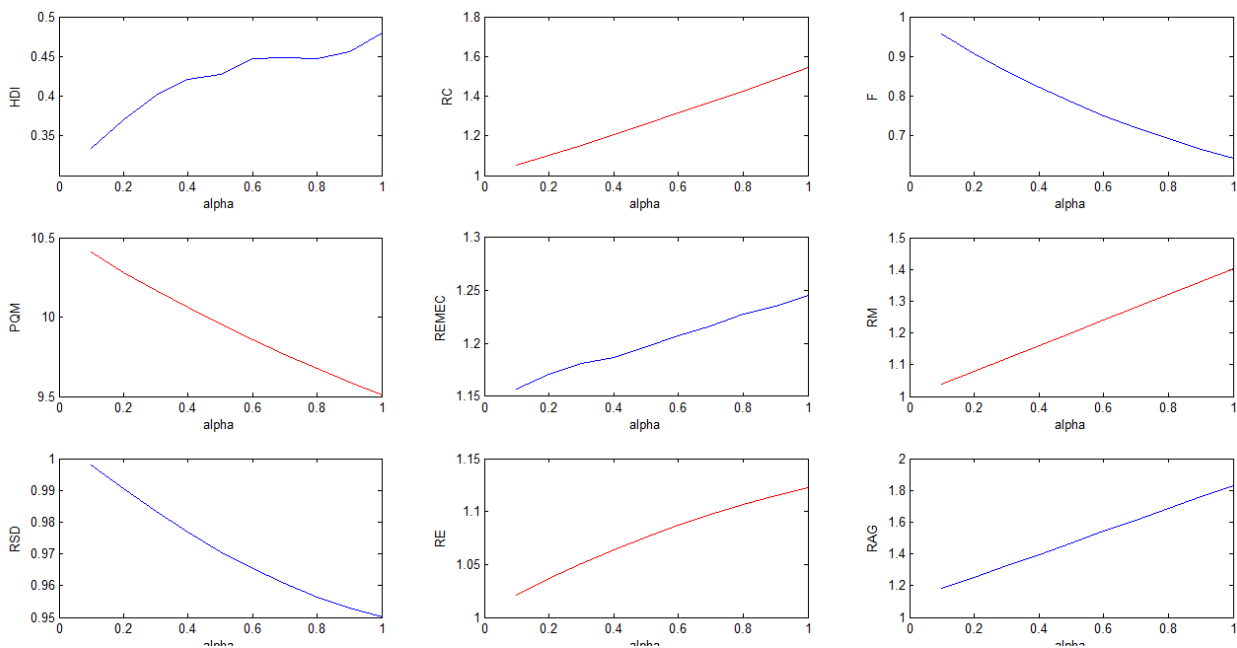

(a)

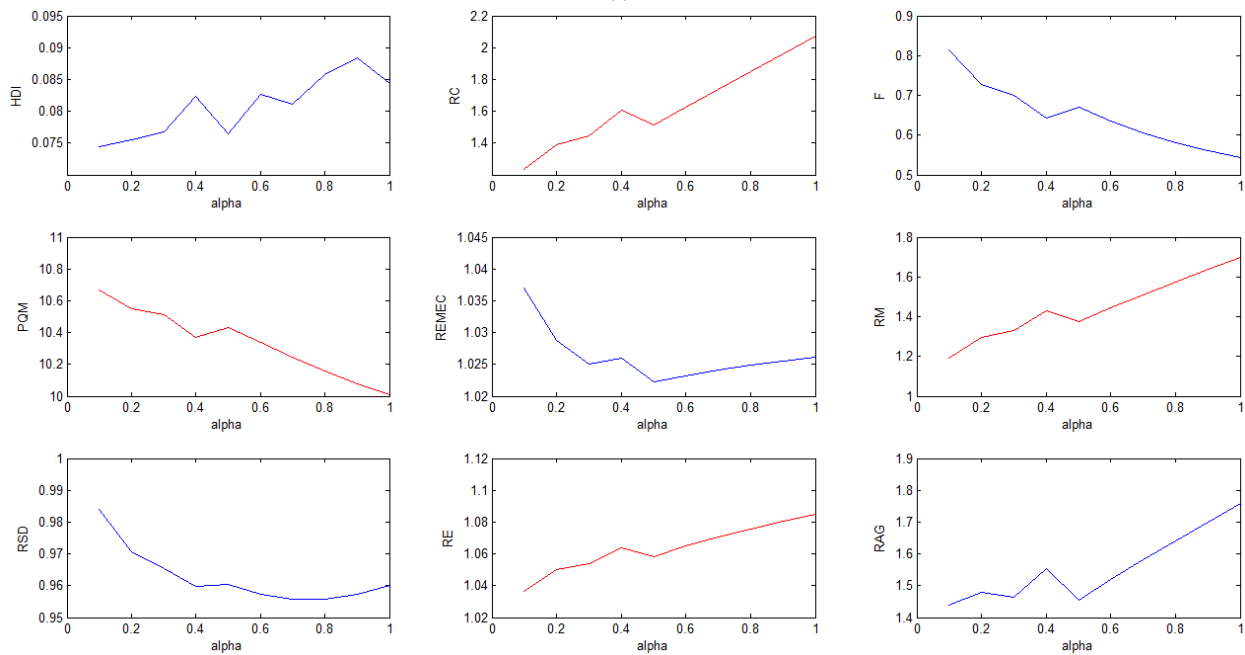

(b)

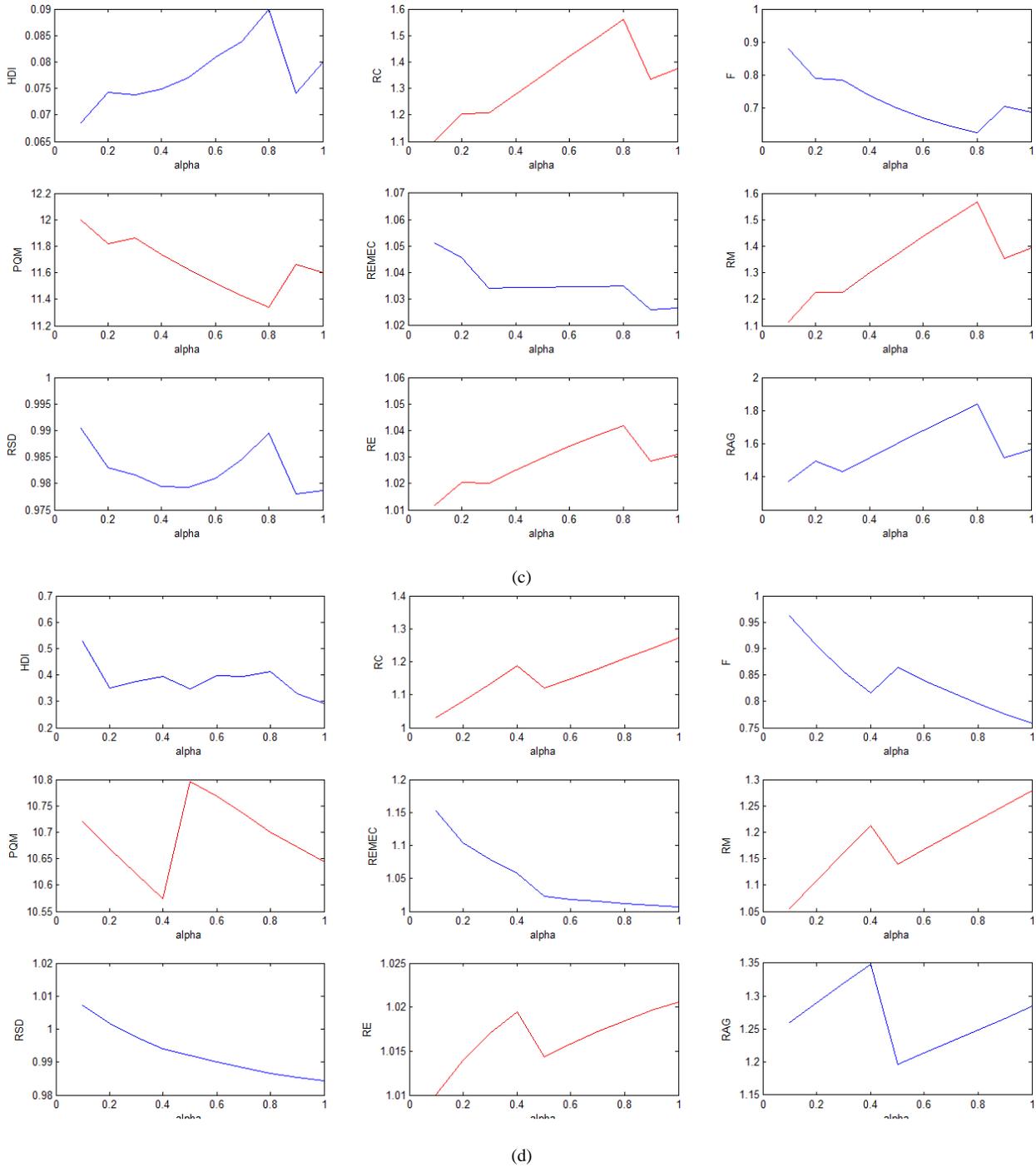

(c)

(d)

**Fig. 3** Plots of HDI, RC, F, PQM, REMEC, RM, RSD, RE and RAG vs α values for the corresponding images in Fig. 5

Results are shown in Table 1A and 1B for various values of α and the additional plots indicate which metrics yield a consistent profile, suitable for utilization in the proposed algorithm to determine optimum operation.

Table1A-Swan image (EMEC_1 = 13.4315) processed with varying α values

(a)    RGB-based version

| α | RC | PQM | REMEC | RM | RE | RAG | HDI | EMEC_2 |
|---|---|---|---|---|---|---|---|---|
| 0.1 | 1.950804 | **9.710533** | **4.062475** | 1.139711 | 1.057033 | 1.584453 | 3.842041 | **54.56515** |
| 0.2 | 2.01747 | 9.633788 | 2.708473 | 1.273946 | 1.075957 | 1.651993 | 5.839153 | 36.37887 |
| 0.3 | 2.156123 | 9.575234 | 2.068722 | 1.377542 | 1.0915 | 1.706633 | **3.480223** | 27.78606 |
| 0.4 | 2.151809 | 9.551409 | 1.783707 | 1.443896 | 1.097811 | 1.72623 | 6.159236 | 23.95787 |
| 0.5 | 2.161868 | 9.550161 | 1.564516 | 1.487658 | 1.100636 | 1.728097 | 8.713196 | 21.01381 |
| 0.6 | 2.263641 | 9.563533 | 1.404743 | 1.513635 | 1.105295 | 1.721371 | 3.522582 | 18.86782 |
| 0.7 | 2.348247 | 9.483993 | 1.304821 | 1.596533 | 1.114682 | 1.792779 | 3.647656 | 17.52571 |
| 0.8 | 2.433833 | 9.411328 | 1.230118 | 1.676993 | 1.122932 | 1.862192 | 3.76033 | 16.52234 |
| 0.9 | **2.754439** | 9.342021 | 1.182487 | **1.130183** | | **1.929413** | 3.866115 | 15.88258 |
| 1.0 | 2.4905 | 9.409997 | 1.195443 | 1.698262 | 1.126215 | 1.852645 | 9.881661 | 16.0566 |

(b)    HSI-based version

| α | RC | F | PQM | REMEC | RM | RE | RAG | HDI | EMEC_2 | n |
|---|---|---|---|---|---|---|---|---|---|---|
| 0.1 | 1.169316 | **0.892509** | 9.718295 | **2.696577** | 1.117712 | 1.059991 | 1.547785 | 0.544388 | **36.2191** | 3 |
| 0.2 | 1.207879 | 0.837854 | 9.777064 | 1.71664 | 1.156832 | 1.067415 | 1.501061 | 0.45767 | 23.05706 | 2 |
| 0.3 | 1.308374 | 0.76954 | 9.619246 | 1.655591 | 1.23233 | 1.088455 | 1.632231 | 0.417717 | 22.23708 | 2 |
| 0.4 | 1.409792 | 0.713669 | 9.474671 | 1.639695 | 1.305785 | 1.10608 | 1.760726 | 0.354899 | 22.02357 | 2 |
| 0.5 | 1.259512 | 0.783899 | **9.813925** | 1.222452 | 1.201332 | 1.076205 | 1.470197 | 0.407836 | 16.41937 | 1 |
| 0.6 | 1.314566 | 0.750194 | 9.721417 | 1.226641 | 1.241433 | 1.087118 | 1.542264 | 0.390349 | 16.47563 | 1 |
| 0.7 | 1.370684 | 0.719442 | 9.633522 | 1.235622 | 1.281696 | 1.097157 | 1.614517 | 0.356126 | 16.59627 | 1 |
| 0.8 | 1.42755 | 0.691512 | 9.549365 | 1.243909 | 1.321999 | 1.106426 | 1.68697 | 0.323162 | 16.70757 | 1 |
| 0.9 | 1.485359 | 0.665995 | 9.469299 | 1.254706 | 1.3622 | 1.114984 | 1.759432 | 0.305127 | 16.85259 | 1 |
| 1.0 | **1.544064** | 0.64293 | 9.390623 | 1.262071 | **1.402677** | **1.122974** | **1.831954** | **0.301517** | 16.95151 | 1 |

Table1B White house image (EMEC_1 = 55.281) processed with varying α values

(a)    RGB-based version

| α | RC | F | PQM | EMEC | RM | RSD | RE | RAG | HDI | EMEC_2 |
|---|---|---|---|---|---|---|---|---|---|---|
| 0.1 | 2.241584 | **0.624063** | **9.600123** | **1.257222** | 1.512764 | 0.971627 | 1.073235 | 2.105557 | **2.660464** | 69.50051 |
| 0.2 | 2.738593 | 0.517312 | 9.555867 | 1.110081 | 1.790538 | 0.962427 | 1.09028 | 2.085729 | 3.49404 | 61.3664 |
| 0.3 | 3.192057 | 0.482354 | 9.500091 | 1.089795 | 1.969116 | 0.974582 | 1.097671 | 2.087205 | 4.30469 | 60.24498 |
| 0.4 | 3.600884 | 0.466439 | 9.42157 | 1.083594 | 2.12165 | 0.994797 | 1.101309 | 2.122015 | 4.879537 | 59.90218 |
| 0.5 | 3.976286 | 0.466008 | 9.360144 | 1.083687 | 2.231235 | 1.019693 | 1.103976 | 2.139111 | 5.543198 | 59.90729 |
| 0.6 | 4.189694 | 0.45592 | 9.275742 | 1.077517 | 2.352242 | 1.035584 | 1.105482 | 2.196076 | 5.529459 | 59.56621 |
| 0.7 | 4.474915 | 0.467031 | 9.245989 | 1.082029 | 2.399147 | 1.058525 | **1.106884** | 2.178177 | 6.198216 | 59.81567 |
| 0.8 | 4.447858 | 0.454577 | 9.203859 | 1.074391 | 2.44737 | 1.05476 | 1.105612 | 2.194681 | 5.870407 | 59.3934 |
| 0.9 | 4.376283 | 0.443248 | 9.169594 | 1.06819 | **2.479031** | 1.048248 | 1.104136 | **2.204662** | 5.444825 | 59.05062 |
| 1.0 | **4.523025** | 0.457684 | 9.187557 | 1.07477 | 2.451382 | **1.059225** | 1.105688 | 2.147726 | 6.112797 | 59.41435 |

(b)HSI-based version

| α | RC | F | PQM | EMEC | RM | RSD | RE | RAG | HDI | EMEC_2 | n |
|---|---|---|---|---|---|---|---|---|---|---|---|
| 0.1 | 1.726359 | 0.652284 | 9.755601 | 1.166088 | 1.503682 | 0.990367 | 1.074309 | 2.230264 | 0.333856 | 64.46253 | 15 |
| 0.2 | 2.205314 | 0.551616 | 9.683107 | 1.068687 | 1.776071 | 0.989802 | 1.093401 | 2.297439 | 0.232257 | 59.07807 | 13 |
| 0.3 | 2.561057 | 0.513961 | 9.589622 | 1.050221 | 1.969099 | 1.006002 | 1.10235 | 2.380234 | 0.190937 | 58.0573 | 12 |
| 0.4 | 2.796342 | 0.499079 | 9.517443 | 1.043883 | 2.094458 | 1.0224 | 1.105847 | 2.425755 | 0.179012 | 57.70689 | 11 |
| 0.5 | 2.948679 | 0.492451 | 9.466611 | 1.040746 | 2.175333 | 1.03501 | 1.107408 | 2.446897 | 0.172828 | 57.5335 | 10 |
| 0.6 | 3.04224 | 0.489006 | 9.427192 | 1.038717 | 2.225463 | 1.043199 | 1.107998 | 2.45046 | 0.169859 | 57.4213 | 9 |
| 0.7 | 3.092515 | 0.486938 | 9.399805 | 1.03716 | 2.25249 | 1.047294 | 1.108129 | 2.440749 | 0.167553 | 57.33525 | 8 |
| 0.8 | 3.223564 | 0.485594 | 9.331406 | 1.036578 | 2.326538 | 1.062898 | 1.10841 | 2.491486 | 0.172049 | 57.30305 | 8 |
| 0.9 | **3.22406** | 0.484498 | 9.319009 | 1.035489 | 2.326349 | 1.061655 | 1.10828 | 2.466307 | 0.170268 | 57.24286 | 7 |
| 1.0 | 3.314023 | 0.484035 | 9.264939 | 1.035103 | 2.380389 | 1.073402 | 1.107922 | 2.50266 | 0.180999 | 57.22155 | 7 |

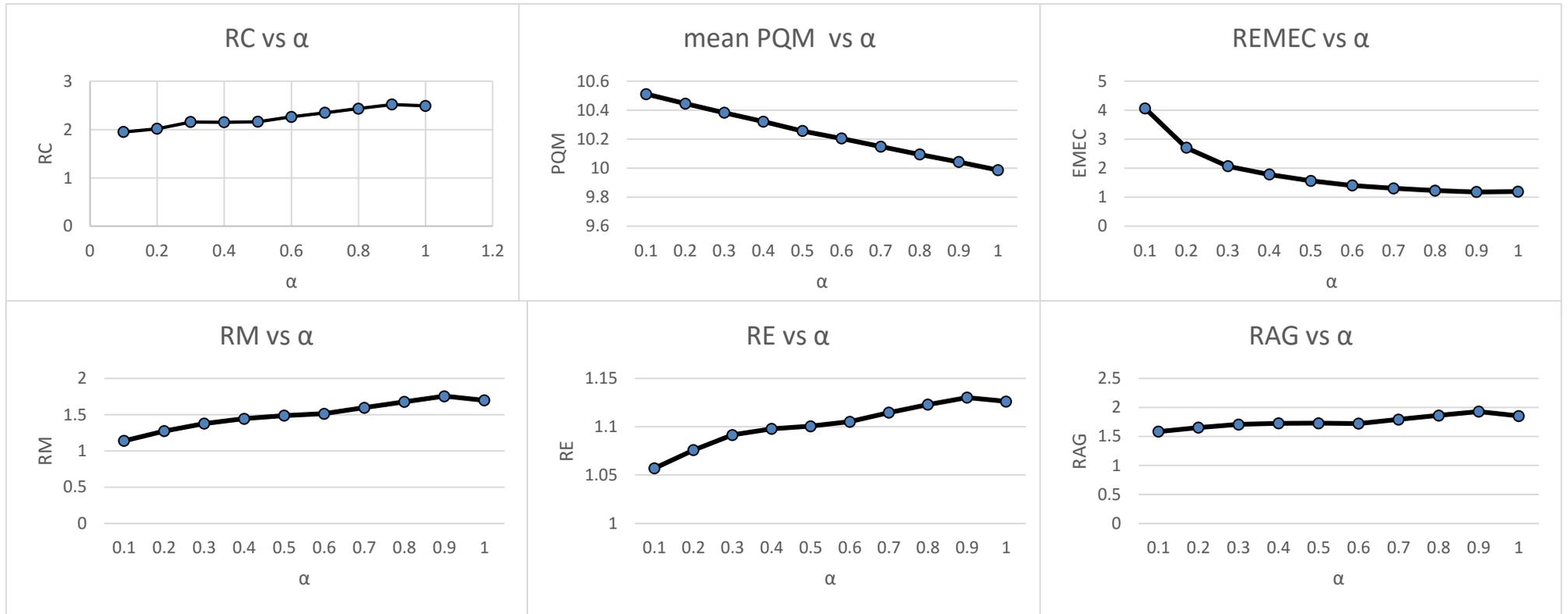

(a)

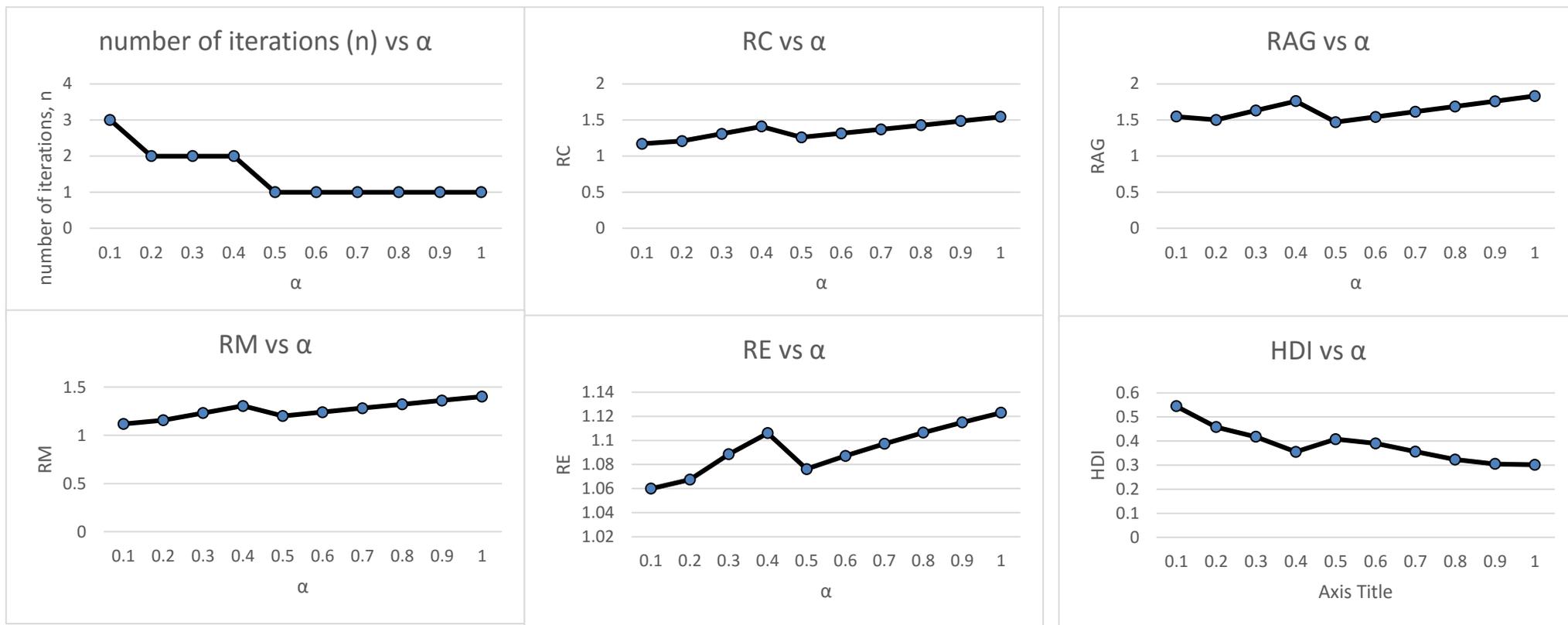

(b)

**Fig. 4** Processed Swan Image at varying values of α against image quality metrics using (a) RGB-IV- and (b) HSI-based PA

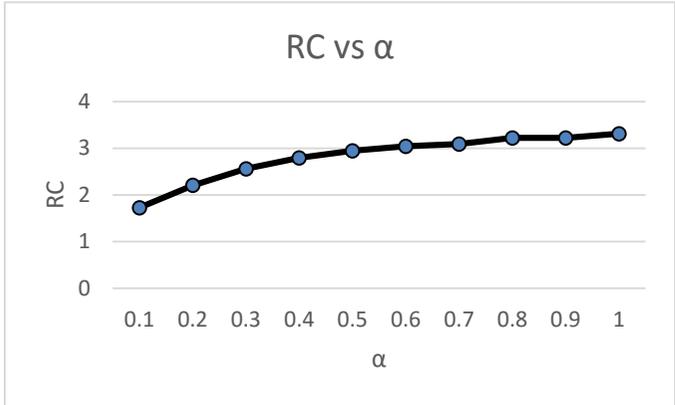

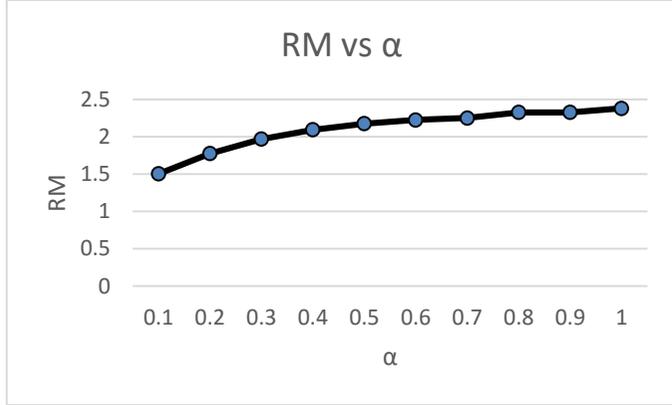

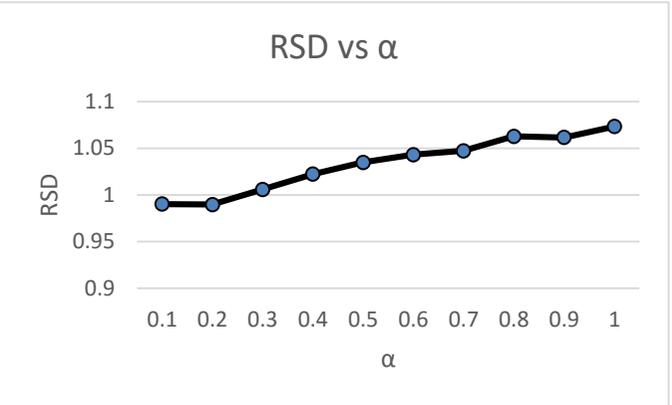

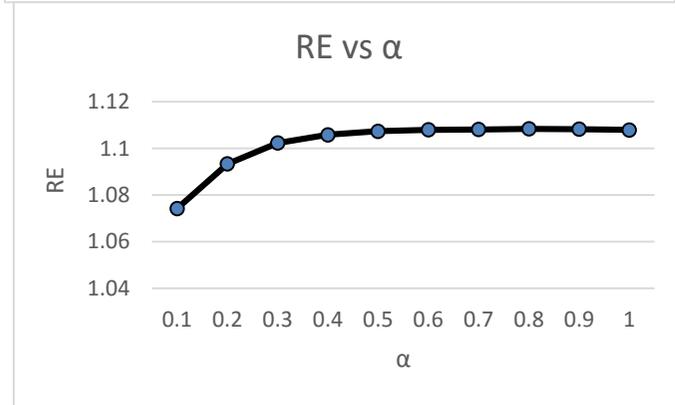

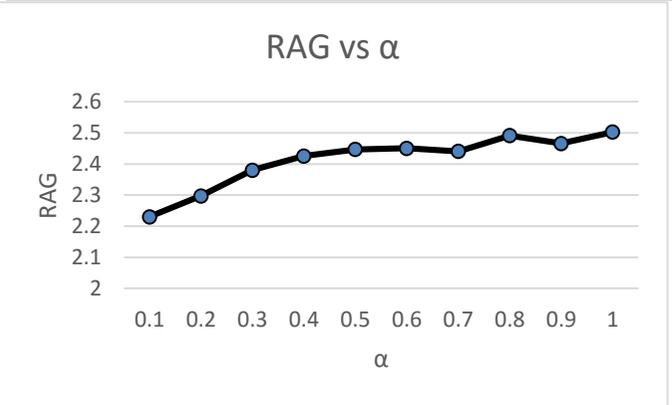

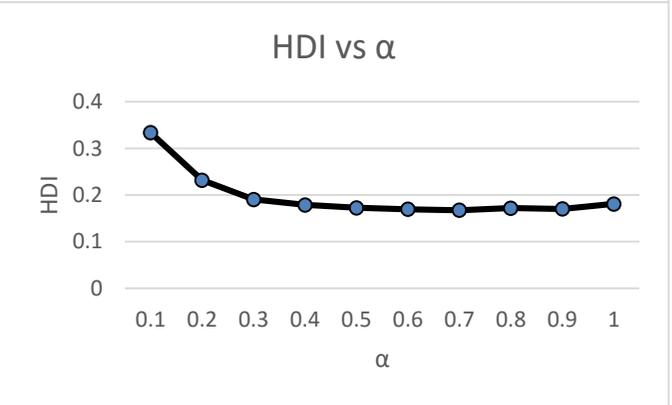

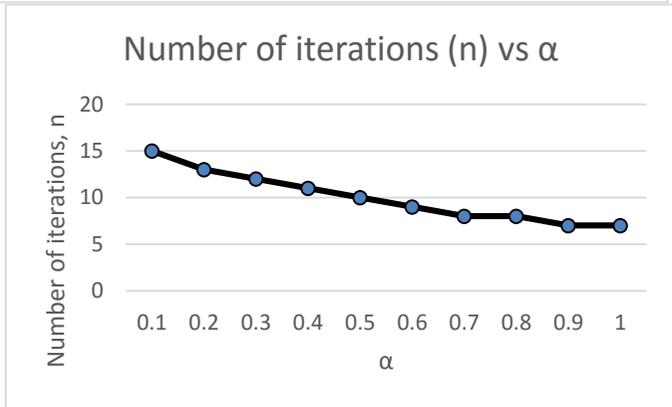

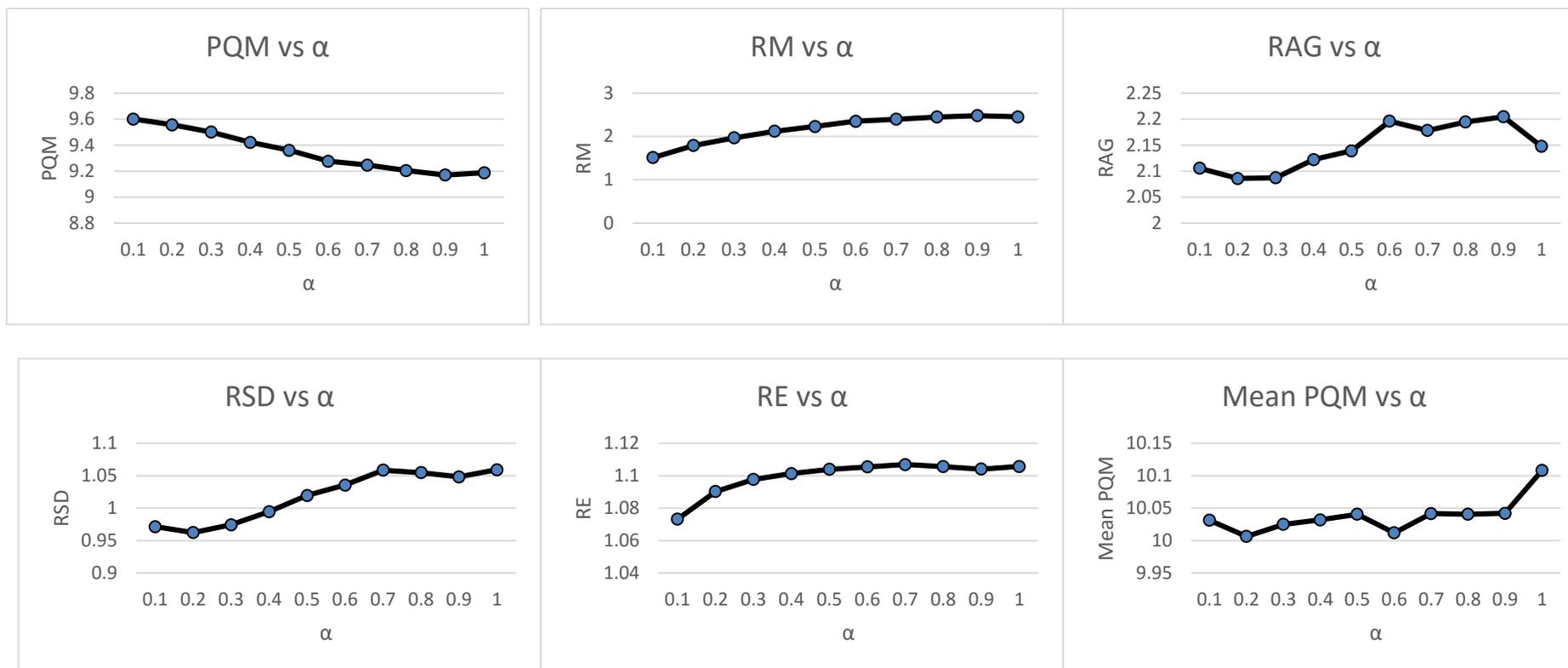

**Fig. 5** Processed White House image at varying values of α against image quality metrics using (a) HSI-PA and (b) RGB-PA

## 4.1 Visual results

We present a sample of results to show how visually striking the results are using the proposed algorithm. Note the preservation of details in bright areas coupled with detail enhancement in formerly dark regions.

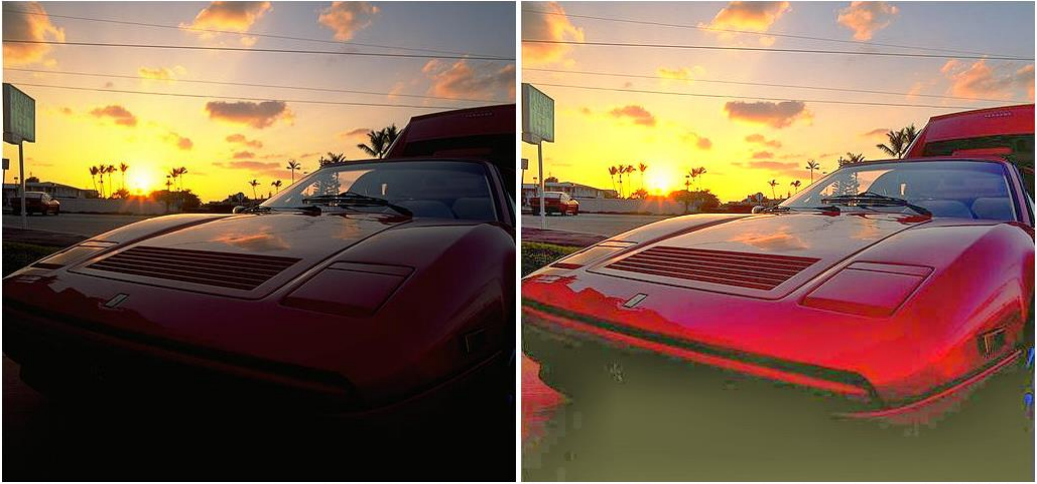

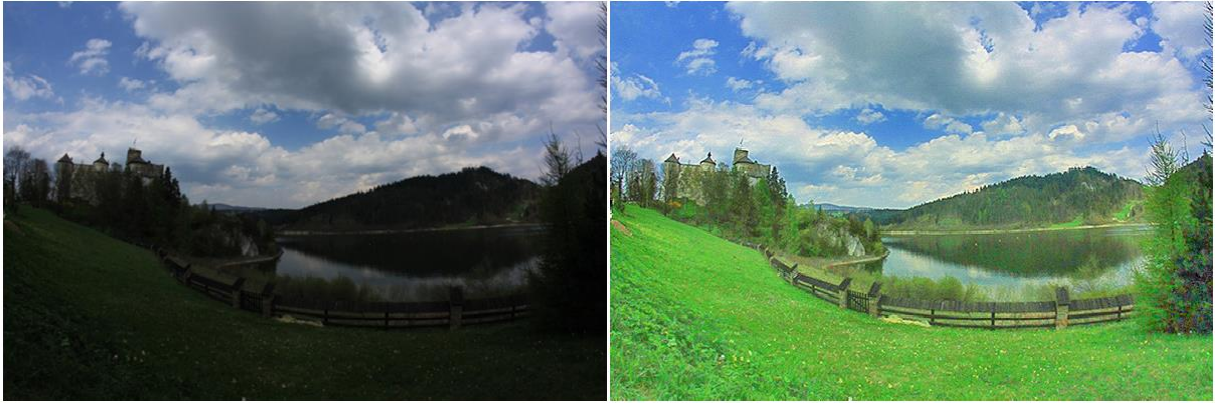

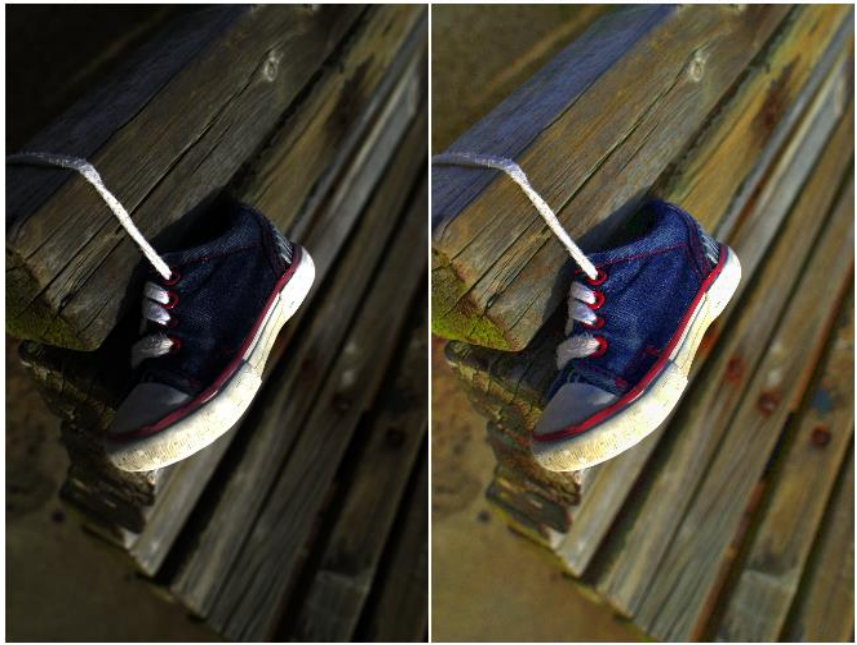

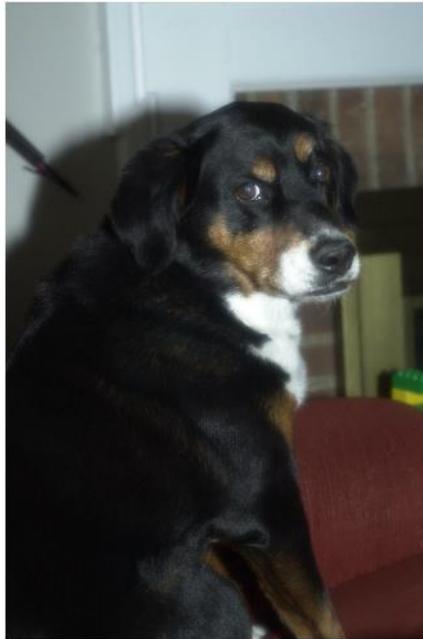 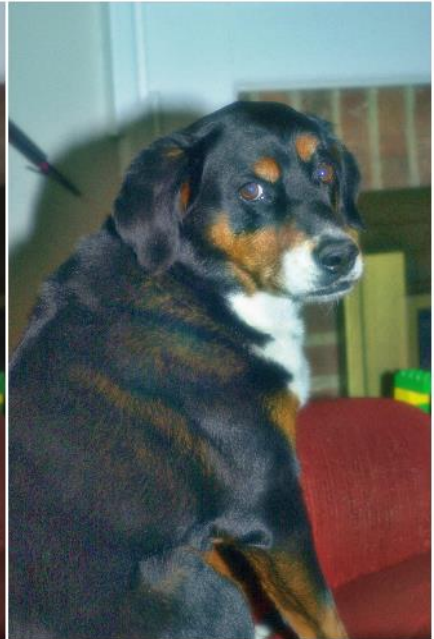

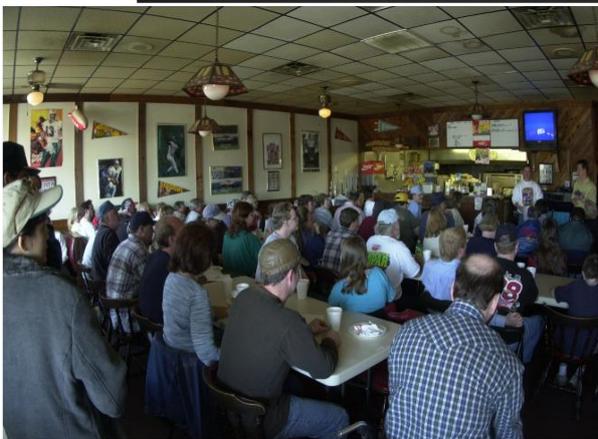 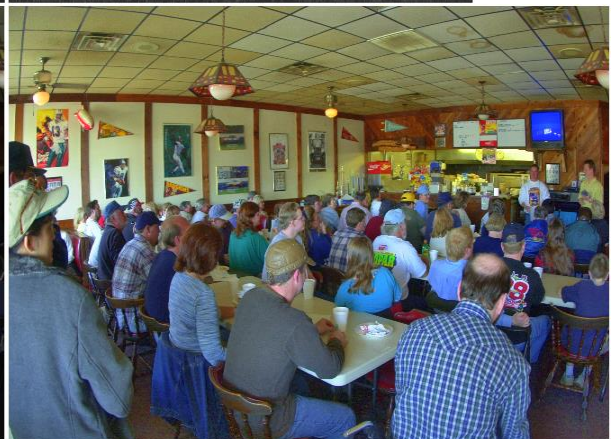

(a)            (b)

**Fig. 6** (a) Original image processed with (b) PA

## 4.2 Comparisons with other algorithms from the literature

Results using PA are compared with those obtained using conventional algorithms from the literature.

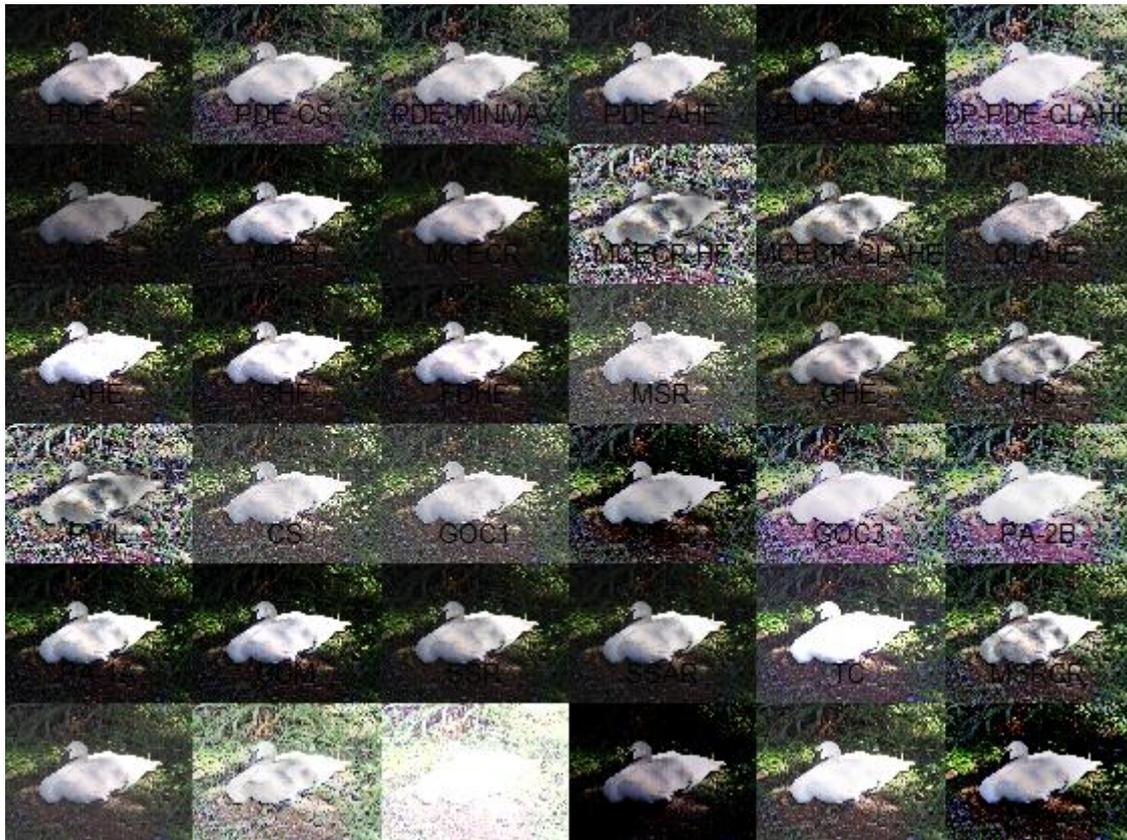

(a)

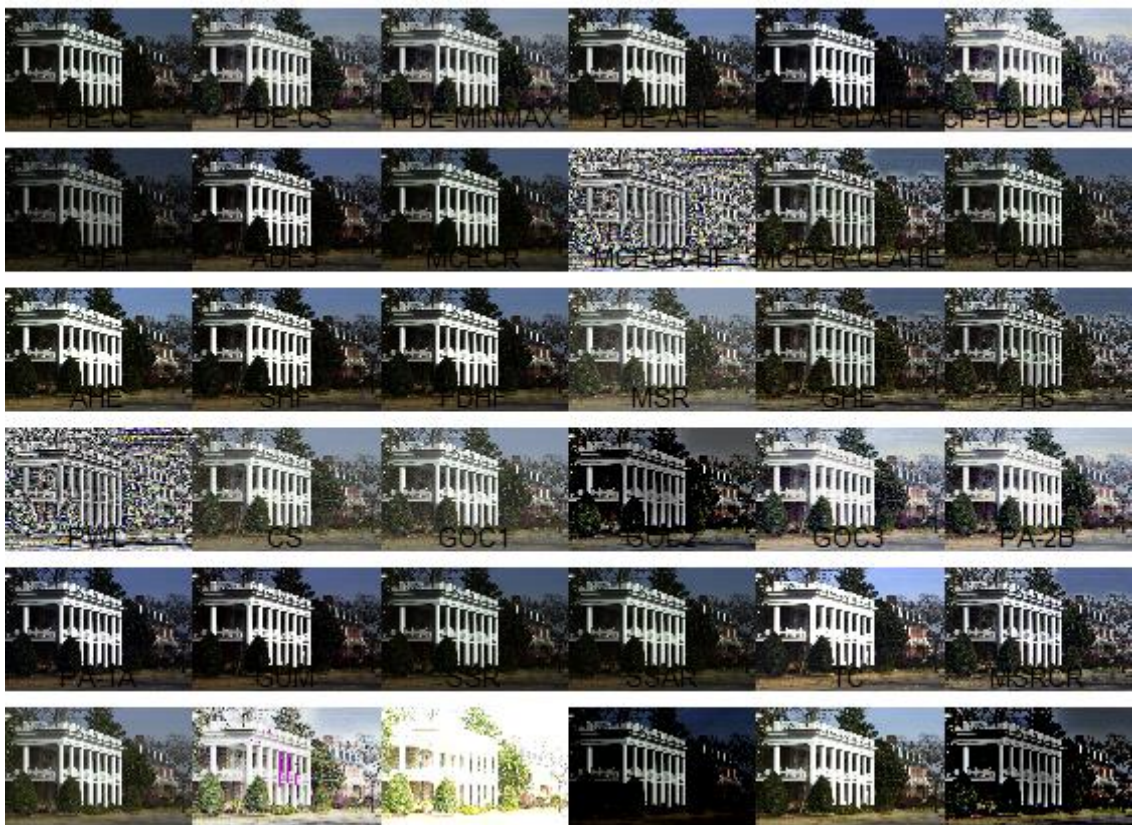

(b)

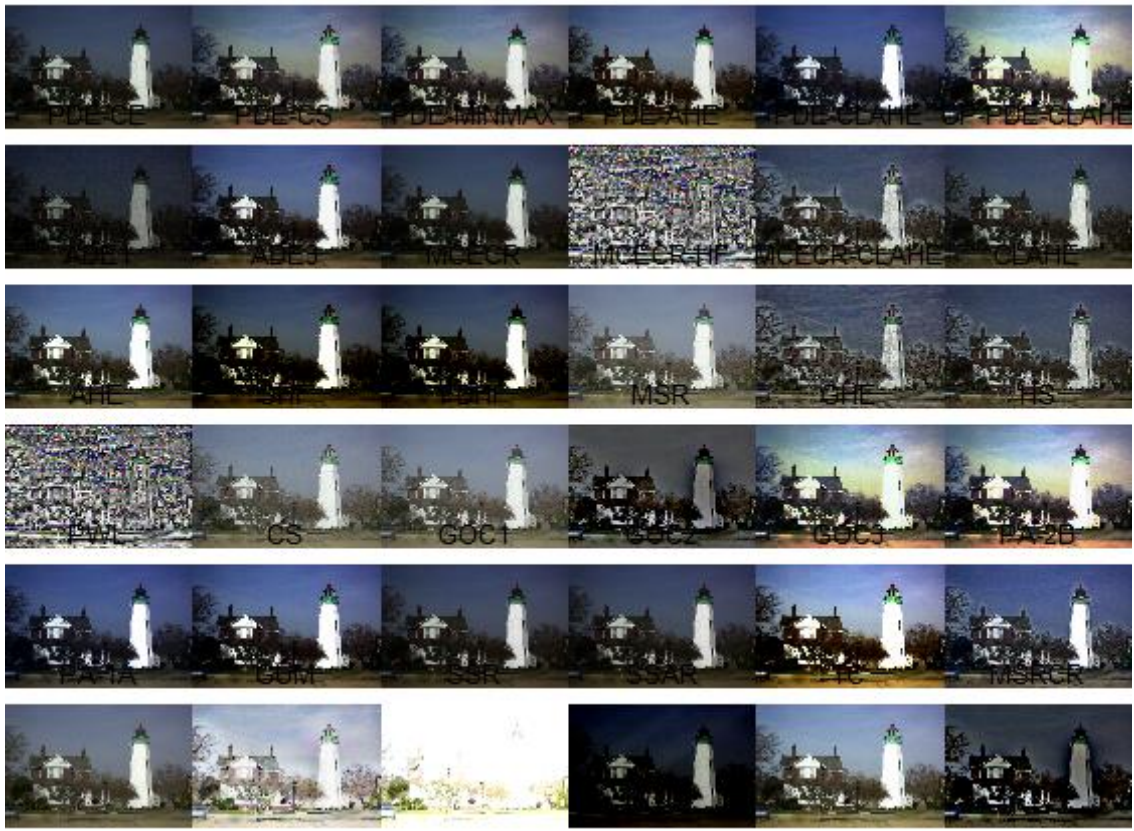

(c)

| Original Image | PDE_HS | PDE_GOC2 | PDE_GOC3 | PDE_PWL | PDE_GHE |
|---|---|---|---|---|---|
| PDE_CE | PDE_CS | PDE_MINMAX | PDE_AHE | PDE_CLAHE | CP_PDE_CLAHE |
| ADE1 | ADE3 | MCECR | MCECR_HF | MCECR_CLAHE | CLAHE |
| AHE | SHF | FDHF | MSR | GHE | HS |
| PWL | CS | GOC1 | GOC2 | GOC3 | RGB-PA |
| HSI-PA | GUM | SSR | SSAR | TC | MSRCR |

(d)

**Fig. 7** image results for processing (a) swan (b) white house (c) light house images processed with various algorithms (d) key to figures

These algorithms include histogram specification (HS) [25], gain offset correction (GOC) and variants [26] (designated as GOC1, GOC2 and GOC3 for ease of notation [27]), piecewise linear transform (PWL) [28], global, adaptive and contrast limited adaptive histogram equalization (GHE, AHE and CLAHE) [1] [8], contrast stretching (CS) [29] [30], Homomorphic filter (SHF & FDHF) [2] and their PDE-based formulations, generalized un-sharp masking (GUM) [31], splitting signal alpha rooting (SSAR) [32], tonal correction (TC) [33], Anisotropic Diffusion-based enhancement (ADE1 & ADE3) [20] and single and multi-scale Retinex with colour restoration (SSR & MSRCR) [3]. The relevant numerical results are presented in Table 2.

Table 2. Comparison of PA with popular algorithms

| Algos\Measures | RC | F | PQM | REMEC | RM | RSD | RE | RAG | HDI | EMEC_2 |
|---|---|---|---|---|---|---|---|---|---|---|
| CLAHE | 1.628192 | 0.593445 | 8.840716 | 1.225813 | 1.533186 | 0.953867 | 1.142119 | 2.278457 | 6.69303 | 16.46451 |
| SHF | 1.36406 | 0.492213 | 8.463246 | 0.946139 | 2.044878 | 1.003252 | 1.131727 | 2.924515 | 7.645274 | 12.70807 |
| FDHF | 1.306712 | 0.447208 | 8.913402 | 0.701319 | 2.061137 | 0.960081 | 1.115542 | 2.326636 | 5.14373 | 9.419773 |
| MSR | 1.648464 | 0.973854 | 8.543244 | | 1.126875 | 1.047574 | 0.960945 | 2.131063 | 9.087314 | |
| GHE | 2.345871 | 0.535439 | 8.085046 | 2.073571 | 2.46279 | 1.148335 | 1.076786 | 3.067447 | 11.31385 | 27.85118 |
| PWL | 1.258974 | 1.402977 | | | 1.013946 | 1.192704 | 0.986951 | 1.21825 | 5.112686 | 138.3362 |
| CS | 1.277531 | 1.419 | 9.959905 | | 1.023272 | 1.204999 | 0.970945 | 1.233004 | 5.314325 | |
| GOC3 | 1.642267 | 0.700361 | 9.175185 | 0.612586 | 2.010868 | 1.186733 | 0.956914 | 1.706666 | 13.69751 | 8.227953 |
| RGB-IV-PA | 1.736804 | 0.821281 | 8.685636 | 1.544765 | 1.483414 | 1.103766 | 1.109074 | 2.341781 | 7.441887 | 20.74852 |
| HSI-PA | 1.045835 | 0.51528 | 9.893684 | 0.574538 | 1.74196 | 0.947416 | 1.063819 | 1.203192 | 2.458478 | 7.716908 |
| GUM | 1.967059 | 0.249363 | 8.3413 | 0.692045 | 3.140091 | 0.884885 | 1.169709 | 2.86375 | 7.819348 | 9.295204 |
| SSR | 1.500898 | 0.055287 | 8.660112 | 0.296004 | 4.499635 | 0.49877 | 0.488084 | 1.500135 | 9.328737 | 3.975786 |
| SSAR | 0.914774 | 1.422905 | 10.10053 | | 0.535115 | 0.872593 | 0.598847 | 0.686228 | 12.88947 | |
| TC | 1.62517 | 0.740843 | 9.147615 | 0.867176 | 1.861492 | 1.17434 | 0.947308 | 1.830121 | 1.139818 | 11.64748 |
| MSRCR | 2.320929 | 1.404061 | 8.303884 | | 0.988585 | 1.178148 | 0.769915 | 1.970034 | 10.10144 | |

(a)

| Algos\Measures | RC | F | PQM | REMEC | RM | RSD | RE | RAG | HDI |
|---|---|---|---|---|---|---|---|---|---|
| CLAHE | 1.182219 | 0.69554 | 9.496405 | 0.216414 | 1.352458 | 0.969891 | 1.058809 | 2.014777 | 6.946697 |
| AHE | 2.177299 | 0.604092 | 7.708619 | 0.639598 | 2.155352 | 1.141066 | 1.119599 | 4.622479 | 21.29402 |
| SHF | 1.089509 | 0.514507 | 9.931548 | 0.171683 | 1.900641 | 0.988885 | 1.077754 | 2.129851 | 3.72423 |
| FDHF | 1.042236 | 0.495105 | 9.71461 | 0.126639 | 1.938911 | 0.979777 | 1.068038 | 1.990646 | 2.758373 |
| MSR | 0.97732 | 0.859922 | 9.336886 | | 0.974533 | 0.915436 | 0.810641 | 1.497496 | 16.51339 |
| GHE | 1.247831 | 0.629135 | 9.359232 | 0.246672 | 2.259811 | 1.192362 | 1.037539 | 1.780163 | 18.0895 |
| HS | 1.264442 | 0.658133 | 9.674896 | 0.47204 | 2.232941 | 1.212259 | 1.121929 | 1.727411 | 17.57141 |
| PWL | 1.360793 | 1.15863 | 10.33854 | 0.826203 | 1.170494 | 1.164547 | 0.980422 | 1.174373 | 18.89695 |
| CS | 1.264952 | 1.22717 | 10.28984 | | 1.112081 | 1.168209 | 0.964119 | 1.167855 | 11.10491 |
| GOC3 | 2.031451 | 0.986923 | 9.656788 | 0.127985 | 1.897476 | 1.368452 | 0.956811 | 1.5444 | 13.02924 |
| RGB-IV-PA | 1.382776 | 0.861439 | 9.318563 | 0.20752 | 1.499366 | 1.136491 | 1.044756 | 2.087882 | 15.69897 |
| HSI-PA | 1.074611 | 0.617489 | 9.82751 | 0.526896 | 1.671483 | 1.015935 | 1.059722 | 1.177929 | 1.501142 |
| GUM | 2.247269 | 0.379875 | 8.820002 | 0.583803 | 2.91394 | 1.052109 | 1.086176 | 2.396891 | 8.215082 |
| SSR | 2.808504 | 0.223848 | 7.978485 | 0.94715 | 3.86448 | 0.930083 | 0.412151 | 1.708954 | 17.03105 |
| SSAR | 0.94224 | 1.39559 | 10.56322 | | 0.50096 | 0.836142 | 0.59088 | 0.685964 | 11.05046 |
| TC | 1.59275 | 0.885017 | 9.715628 | 0.998348 | 1.880559 | 1.290088 | 0.949236 | 1.561191 | 1.638593 |
| MSRCR | 1.377766 | 1.087737 | 9.162248 | | 0.839034 | 0.955326 | 0.728938 | 1.463395 | 14.24293 |

(b)

| Algos\Measures | RC | F | PQM | REMEC | RM | RSD | RE | RAG | HDI | EMEC_2 |
|---|---|---|---|---|---|---|---|---|---|---|
| CLAHE | 1.052985 | 0.754813 | 10.47959 | 0.569747 | 1.278047 | 0.982184 | 1.025218 | 2.529385 | 4.193548 | 7.529527 |
| AHE | 2.513004 | 1.094964 | 7.272155 | 2.372797 | 1.98265 | 1.473408 | 1.135379 | 11.52049 | 21.43365 | 31.35783 |
| SHF | 1.005869 | 0.492258 | 11.09777 | 0.358905 | 1.995833 | 0.991194 | 1.042932 | 2.28239 | 1.799167 | 4.74313 |
| FDHF | 0.981311 | 0.472941 | 11.09057 | 0.296963 | 2.02659 | 0.979008 | 1.036979 | 2.038332 | 1.29827 | 3.924537 |
| MSR | 0.715609 | 0.770558 | 10.837 | | 1.061574 | 0.904436 | 0.890535 | 1.911216 | 10.69133 | |
| GHE | 2.083795 | 1.285687 | 10.10695 | 0.62809 | 2.165003 | 1.668388 | 1.041503 | 2.397693 | 15.25921 | 8.300552 |
| HS | 2.157094 | 1.339009 | 10.69543 | | 2.19393 | 1.713969 | 1.146959 | 2.233514 | 16.40375 | |
| PWL | 2.033203 | 1.38031 | 11.65116 | 2.208118 | 1.25649 | 1.316945 | 1.00161 | 1.302065 | 14.68967 | 29.1815 |
| CS | 2.063642 | 1.519461 | 11.11876 | | 1.432111 | 1.475139 | 0.997335 | 1.603608 | 6.933885 | |
| GOC3 | 2.542983 | 1.747642 | 10.24128 | | 1.946965 | 1.844613 | 0.941684 | 2.200659 | 6.332405 | |
| RGB-IV-PA | 1.926655 | 1.0477 | 10.32308 | 0.656215 | 1.485884 | 1.247702 | 1.045428 | 2.611743 | 12.18016 | 8.672245 |
| HSI-PA | 0.959284 | 0.6485 | 11.14831 | 0.886318 | 1.692128 | 1.047542 | 1.039901 | 1.058668 | 1.560984 | 11.71319 |
| GUM | 1.242182 | 0.365995 | 10.16872 | 0.547914 | 3.060389 | 1.058342 | 1.052511 | 2.618102 | 5.750521 | 7.240984 |
| SSR | 1.502208 | 0.1538 | 9.00087 | 0.6366 | 4.174329 | 0.801257 | 0.246309 | 1.129247 | 24.46123 | 8.413018 |
| SSAR | 0.874878 | 1.840887 | 12.71323 | | 0.363358 | 0.817864 | 0.667756 | 0.625645 | 8.442299 | |
| TC | 1.650232 | 0.920235 | 10.99044 | 0.926302 | 2.11531 | 1.3952 | 0.977322 | 1.737885 | 0.94794 | 12.2416 |
| MSRCR | 0.94133 | 1.09185 | 10.65512 | | 0.813566 | 0.942493 | 0.851054 | 1.907999 | 7.982498 | |

(c)

We also test images from the Kodak dataset and the results are shown in Fig. 8.

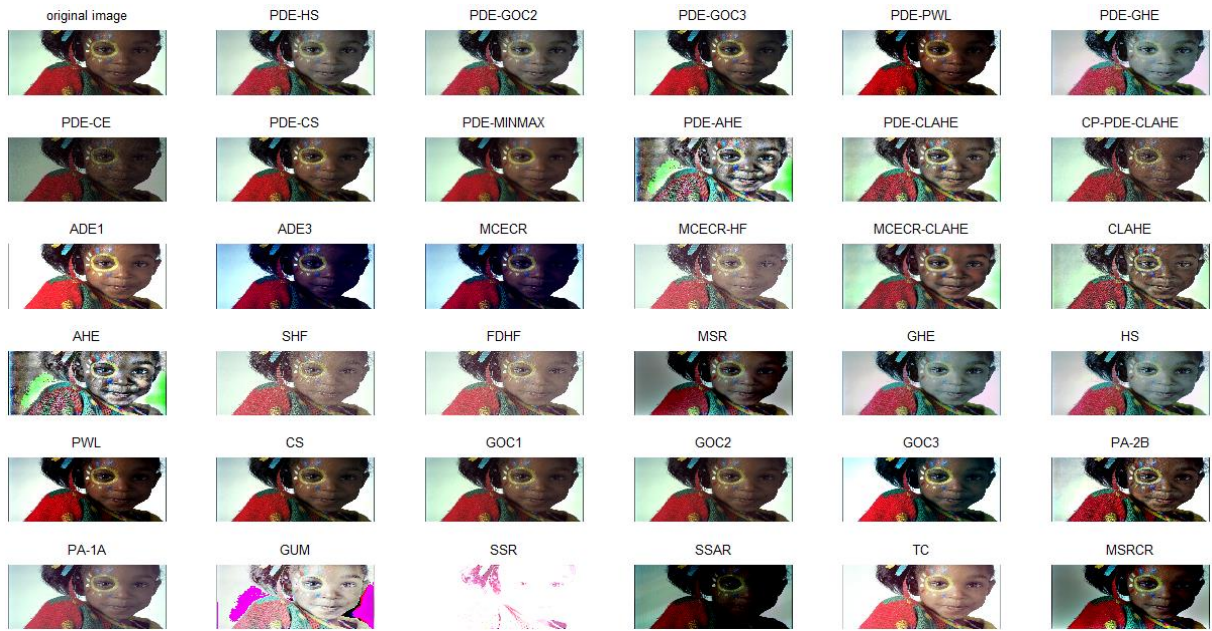

(a)

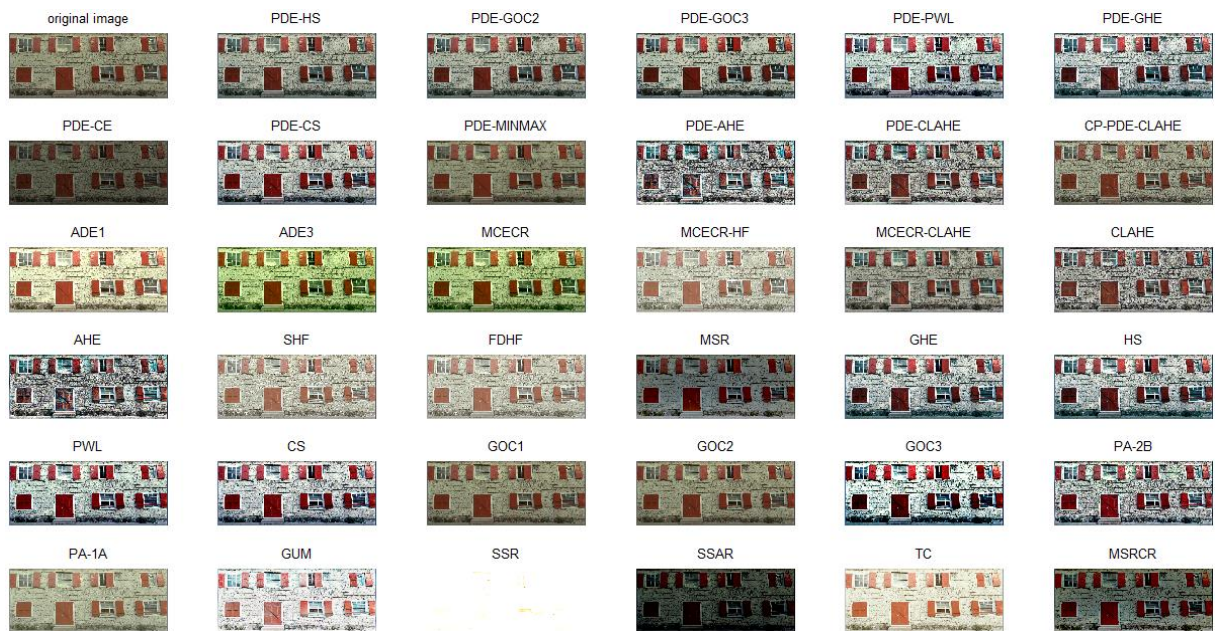

(b)

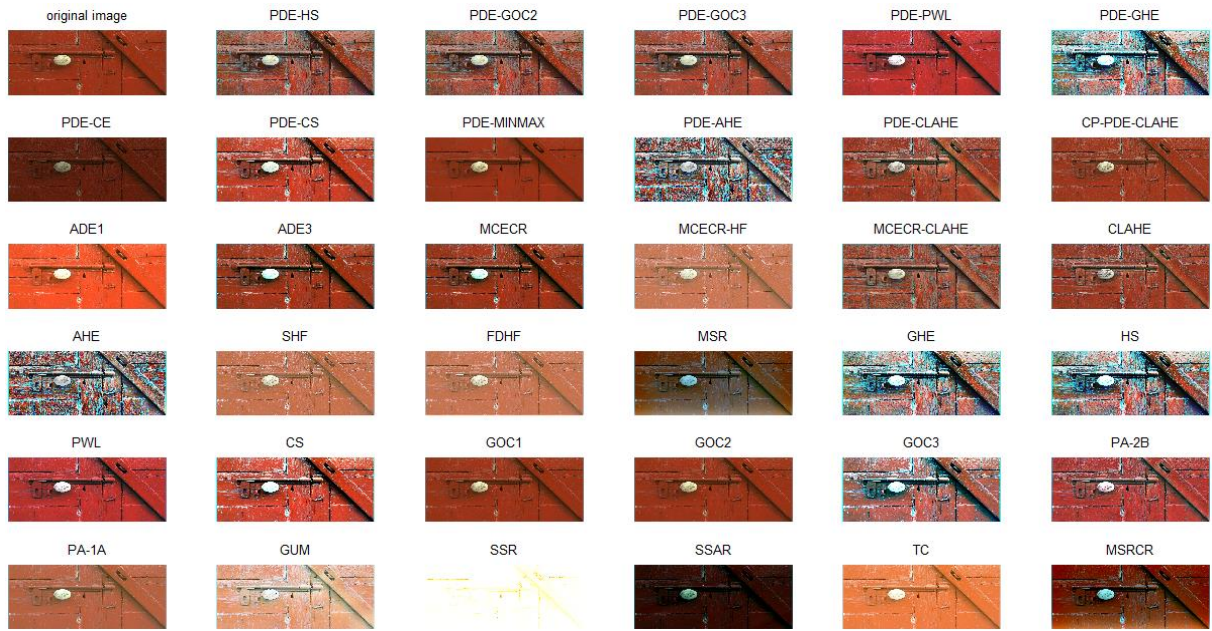

(c)

**Fig. 8** image results for processing (a) girl (b) house (c) bolt images processed with various algorithms

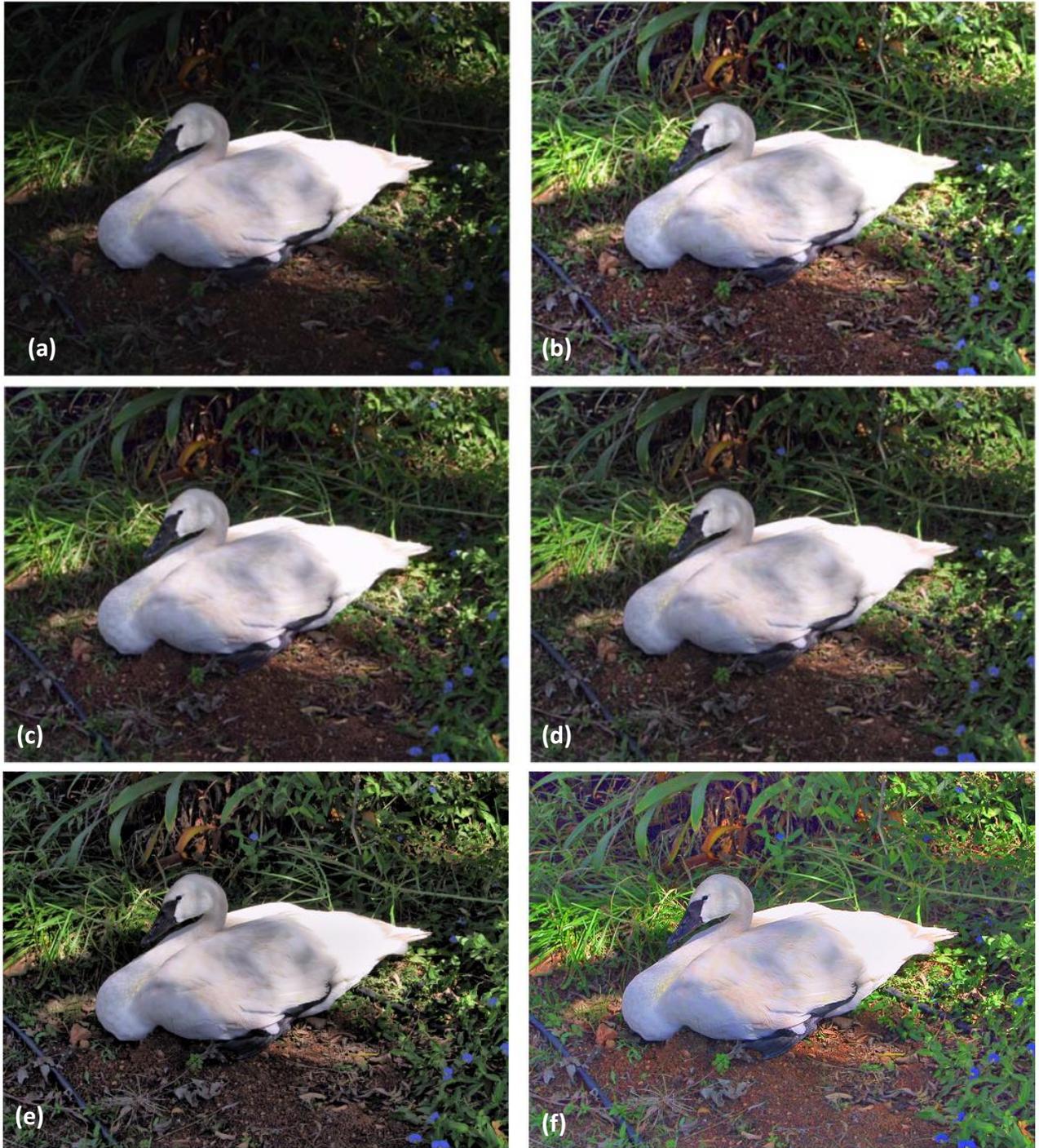

**Fig. 9** (a) Original swan image (b), (c) & (d) results from [34] (e) HSI-PA (f) RGB-PA

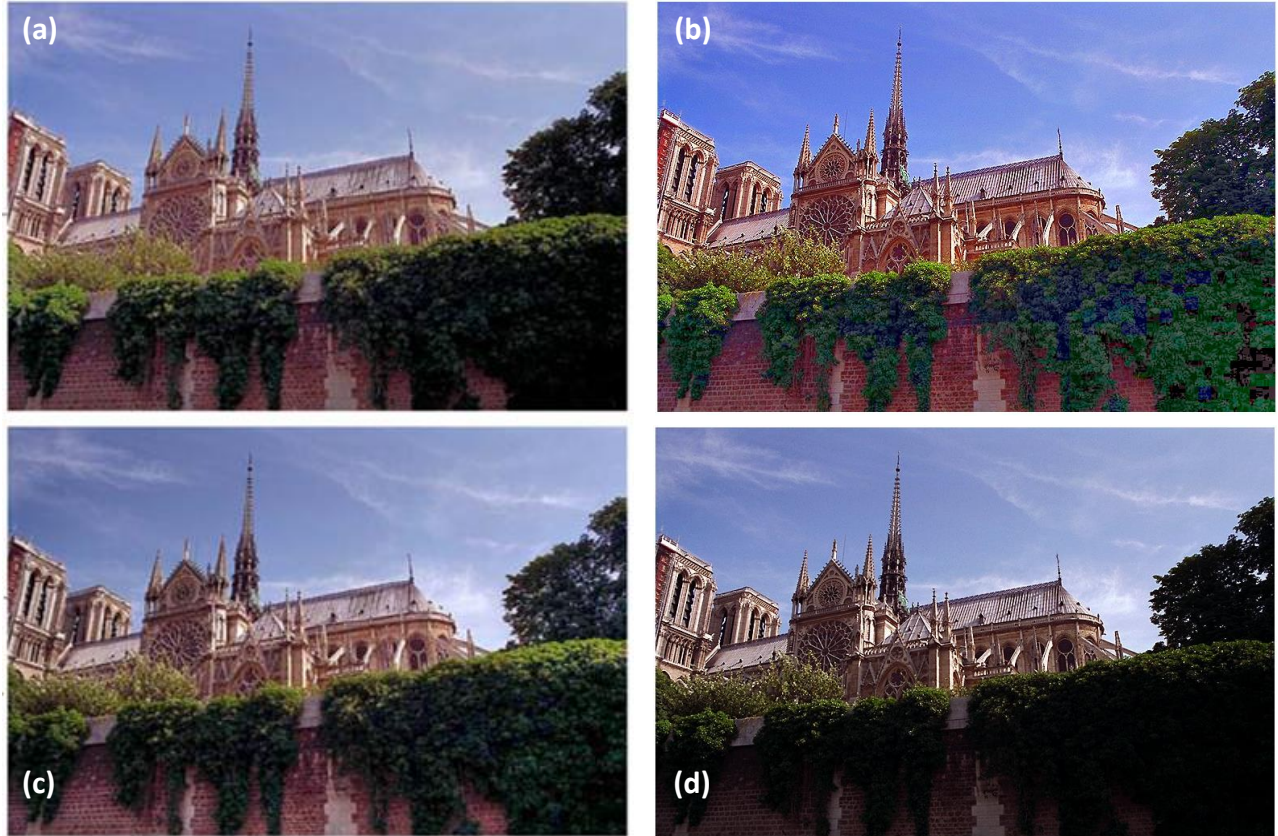

**Fig. 10** (a) & (c) results from [34] (b) RGB-PA (d) original cathedral image

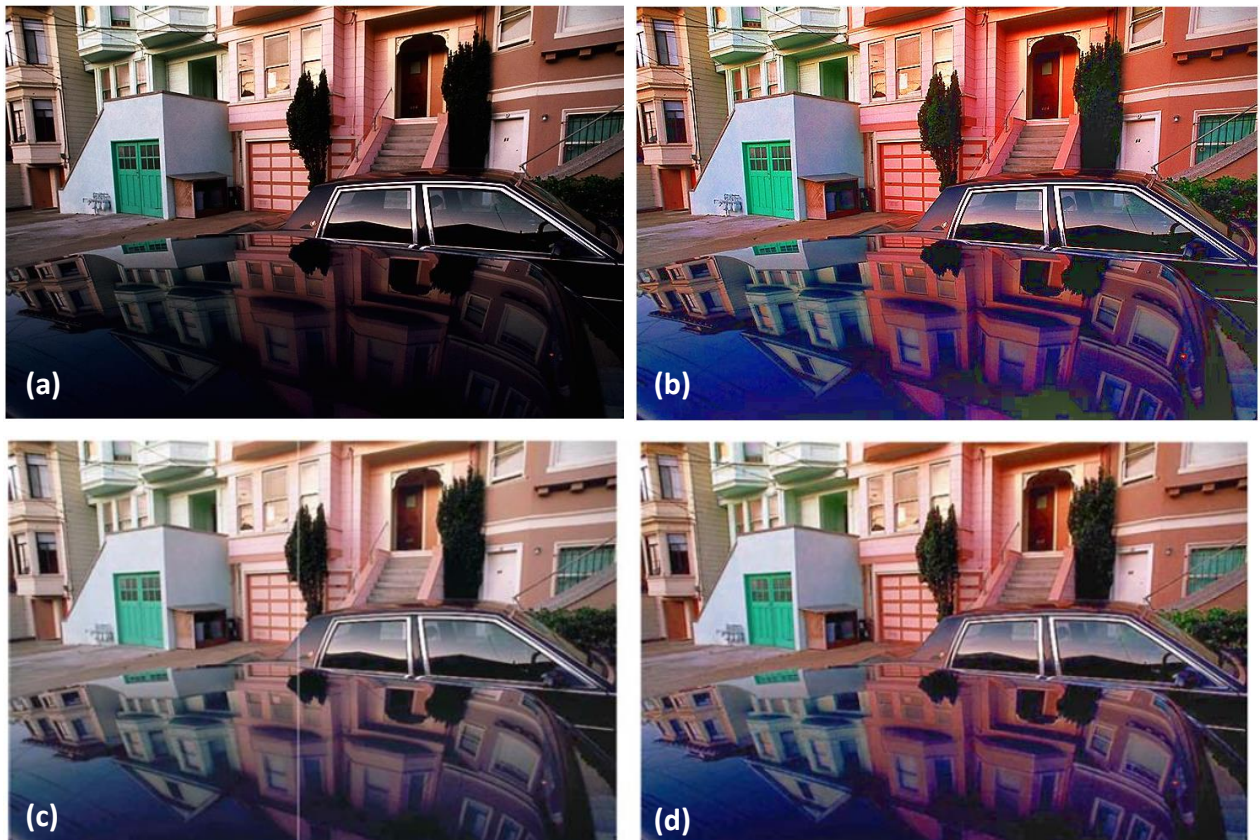

**Fig. 11** (a) original car image (b) RGB-PA (c) & (d) results from [34]

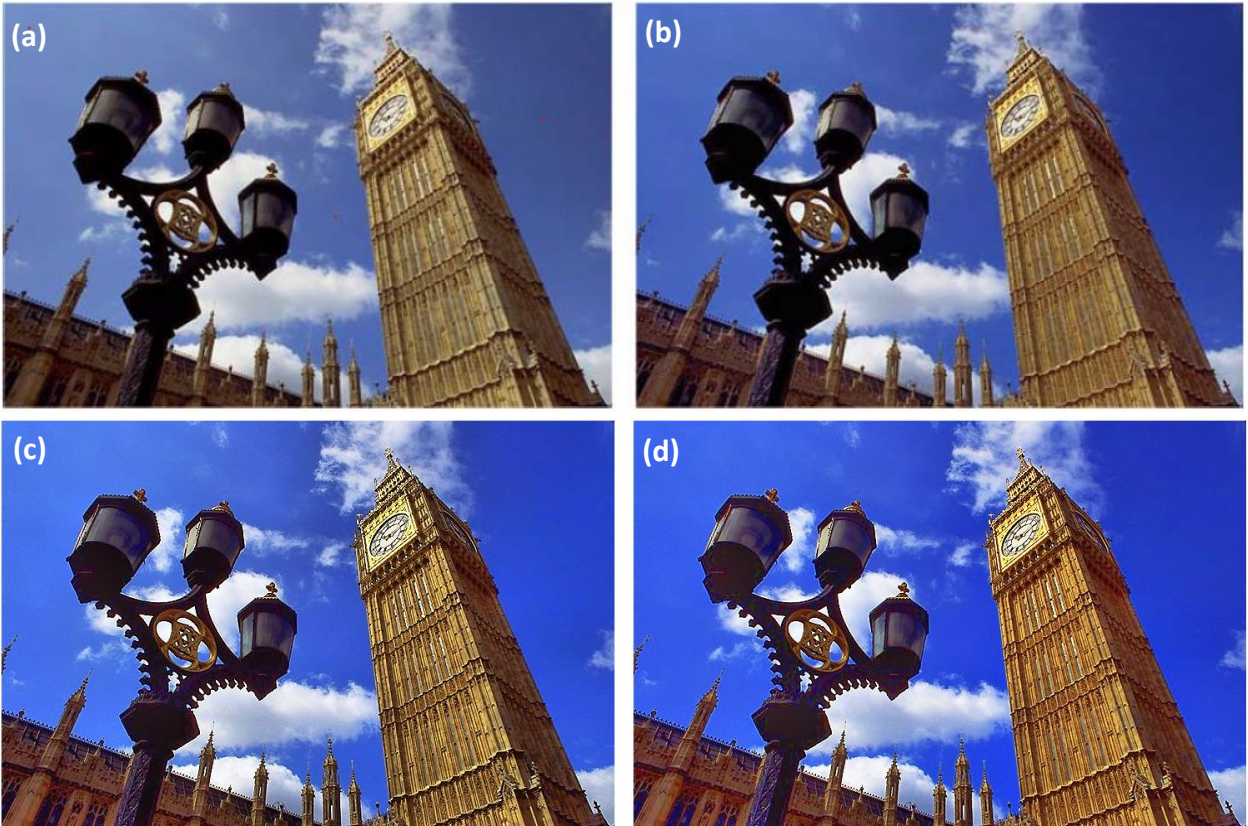

**Fig. 12** (a) & (b) results from [34] (c) HSI- (d) RGB-PA for processed big ben image

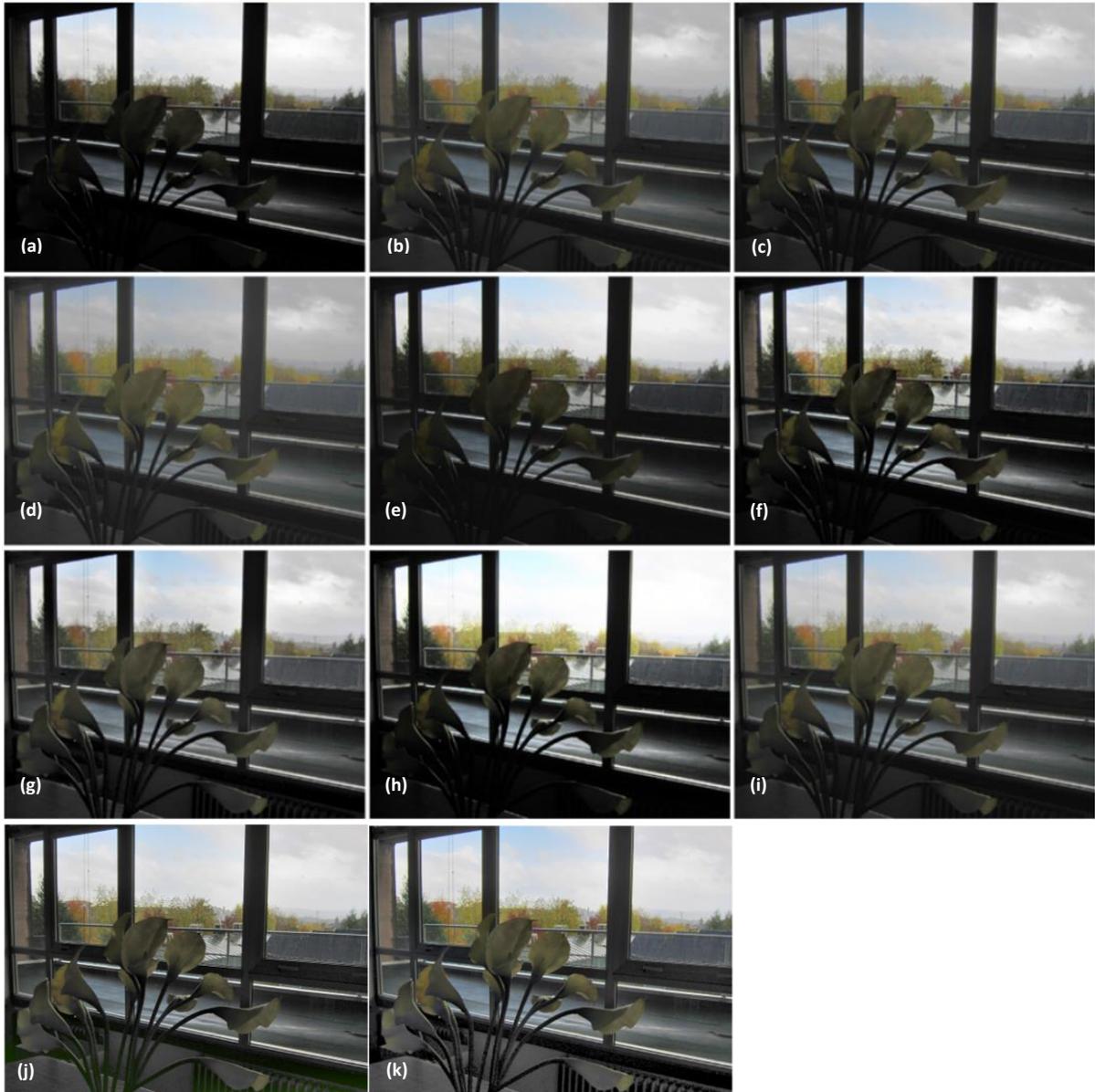

**Fig. 13** Amended figure from [35] showing (a) original iris image (b) to (d) SLIP using various values (e) LCS (f) CLAHE (g) LCC (h) AGC (i) PLR (j) HSI-PA (k) RGB-PA

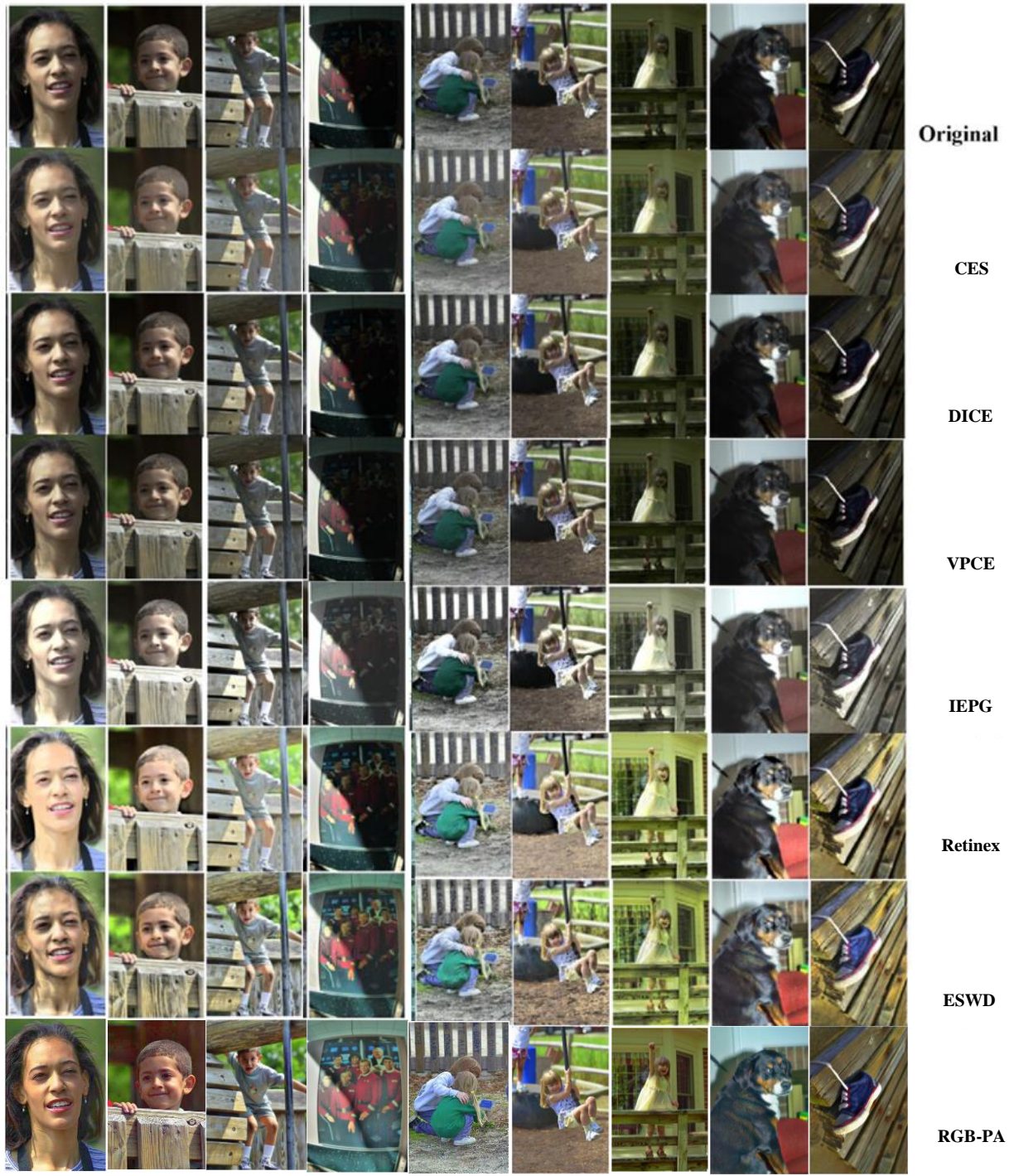

**Fig. 14** Figure from [36] amended with the results from RGB-PA (last row) for visual comparison

Table 3. Comparison of PA with popular algorithms based on (a) CEF and (b) GMSD values

(a)

| Pics\Algorithms | CES [37] | DICE [38] | VPCE [39] | IEGP [40] | MSR [41] | ESWD [36] | PA |
|---|---|---|---|---|---|---|---|
| pic2 (boy & tyre) | 1.117 | 1.018 | 0.936 | 0.853 | 1.893 | 1.903 | **1.96755** |
| pic3 (woman's face) | 1.077 | 1.005 | 0.913 | 0.87 | 1.706 | 1.586 | **1.80695** |
| pic4 (boy & fence) | 1.163 | 1.013 | 0.935 | 0.931 | 2.024 | 2.027 | **2.02823** |
| pic5 (shadow boy) | 1.157 | 1.015 | 0.928 | 0.95 | 1.982 | 2.04 | **2.11065** |
| pic9 (star trek crew) | 1.201 | 1.016 | 0.929 | 0.914 | 2.016 | **2.443** | 2.32648 |
| pic11 (two children) | 1.263 | 1.013 | 0.915 | 0.925 | 1.84 | 1.952 | **2.1209** |
| pic12 (girl & swing) | 1.147 | 1.009 | 0.92 | 0.985 | 1.595 | 1.651 | **1.83339** |
| pic13 (girl standing) | 1.274 | 1.028 | 0.958 | 0.938 | 2.619 | 3.028 | **3.3629** |
| pic14 (girl in car) | 1.196 | 1.006 | 0.936 | 0.862 | 1.733 | 1.555 | **1.9807** |
| pic15 (dog) | 1.296 | 1 | 0.908 | 0.96 | **2.54** | 2.497 | 1.90315 |
| pic16 (surfers) | 1.402 | 1.031 | 0.941 | 0.984 | 2.714 | **3.039** | 2.41513 |
| pic18 (meeting) | 1.134 | 1.018 | 0.933 | 0.884 | 1.89 | **1.999** | 1.79153 |
| pic19 (shoe) | 1.364 | 1.024 | 0.958 | 0.958 | 2.715 | 3.591 | **3.5946** |
| pic20 (house tower) | 1.31 | 1.014 | 0.92 | 1.002 | 2.614 | **2.553** | 2.28488 |
| pic21 (white house) | 1.278 | 1.027 | 0.922 | 0.998 | 2.445 | **2.909** | 2.19273 |

(b)

| Pics\Algorithms | CES [37] | DICE [38] | VPCE [39] | IEGP [40] | MSR [41] | ESWD [36] | PA |
|---|---|---|---|---|---|---|---|
| pic2 (boy & tyre) | 0.044 | 0.038 | 0.004 | 0.096 | 0.153 | 0.119 | 0.0571 |
| pic3 (woman's face) | 0.04 | 0.04 | 0.015 | 0.083 | 0.174 | 0.144 | 0.0941 |
| pic4 (boy & fence) | 0.05 | 0.048 | 0.023 | 0.1 | 0.115 | 0.125 | 0.0732 |
| pic5 (shadow boy) | 0.048 | 0.048 | 0.022 | 0.1 | 0.12 | 0.147 | 0.0572 |
| pic9 (star trek crew) | 0.033 | 0.054 | 0.017 | 0.09 | 0.141 | 0.164 | 0.1143 |
| pic11 (two children) | 0.049 | 0.045 | 0.025 | 0.083 | 0.153 | 0.145 | 0.0687 |
| pic12 (girl & swing) | 0.05 | 0.035 | 0.016 | 0.113 | 0.125 | 0.152 | 0.0201 |
| pic13 (girl standing) | 0.034 | 0.04 | 0.018 | 0.115 | 0.163 | 0.167 | 0.0965 |
| pic14 (girl in car) | 0.042 | 0.053 | 0.004 | 0.105 | 0.186 | 0.19 | 0.1282 |
| pic15 (dog) | 0.034 | 0.044 | 0.007 | 0.07 | 0.138 | 0.138 | 0.0851 |
| pic16 (surfers) | 0.045 | 0.044 | 0.001 | 0.105 | 0.171 | 0.171 | 0.0833 |
| pic18 (meeting) | 0.054 | 0.063 | 0.004 | 0.08 | 0.192 | 0.145 | 0.0536 |
| pic19 (shoe) | 0.033 | 0.018 | 0.013 | 0.125 | 0.16 | 0.156 | 0.1282 |
| pic20 (house tower) | 0.028 | 0.042 | 0.002 | 0.105 | 0.161 | 0.162 | 0.0601 |
| pic21 (white house) | 0.042 | 0.077 | 0.003 | 0.141 | 0.18 | 0.177 | 0.1053 |

## 5. Other possible application areas

The proposed algorithm can also be applied to natural and underwater images with a dominant colour cast. The results are shown in Fig. 15.

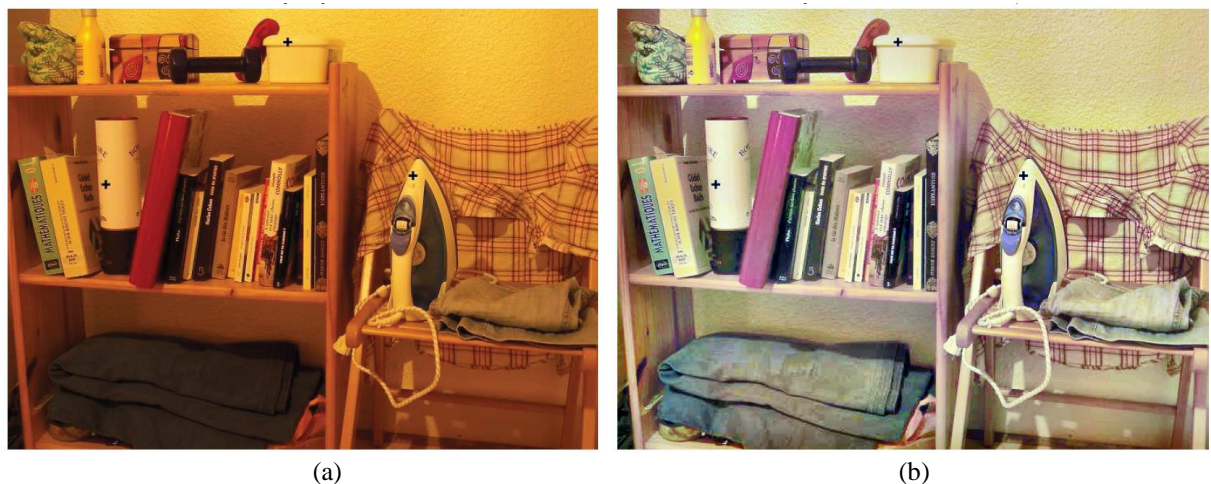

(a)                                           (b)

**Fig. 15** (a) Original colour distorted image (b) Colour and illumination corrected image using PA

The proposed approach can also be used for underwater image enhancement as shown in Fig. 16, though there is greyish tint in some of the images. Additionally, the algorithm can also be used to enhance some hazy images as shown in Fig. 17.

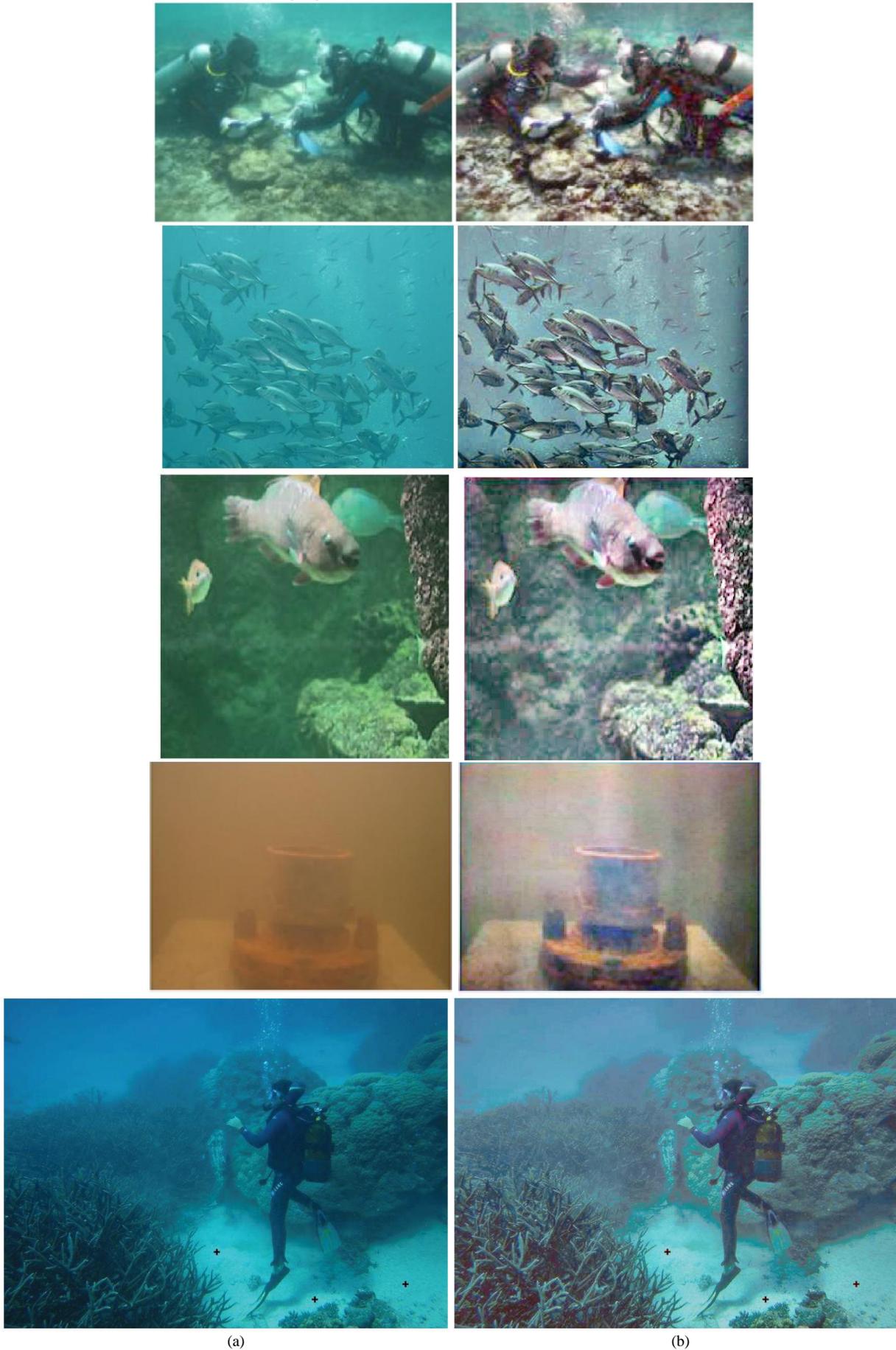

|   |   |
|---|---|
| (a) | (b) |

**Fig. 16** (a) Original colour distorted image (b) Colour and illumination corrected image using PA without RGB-IV or HSI/HSV colour solution

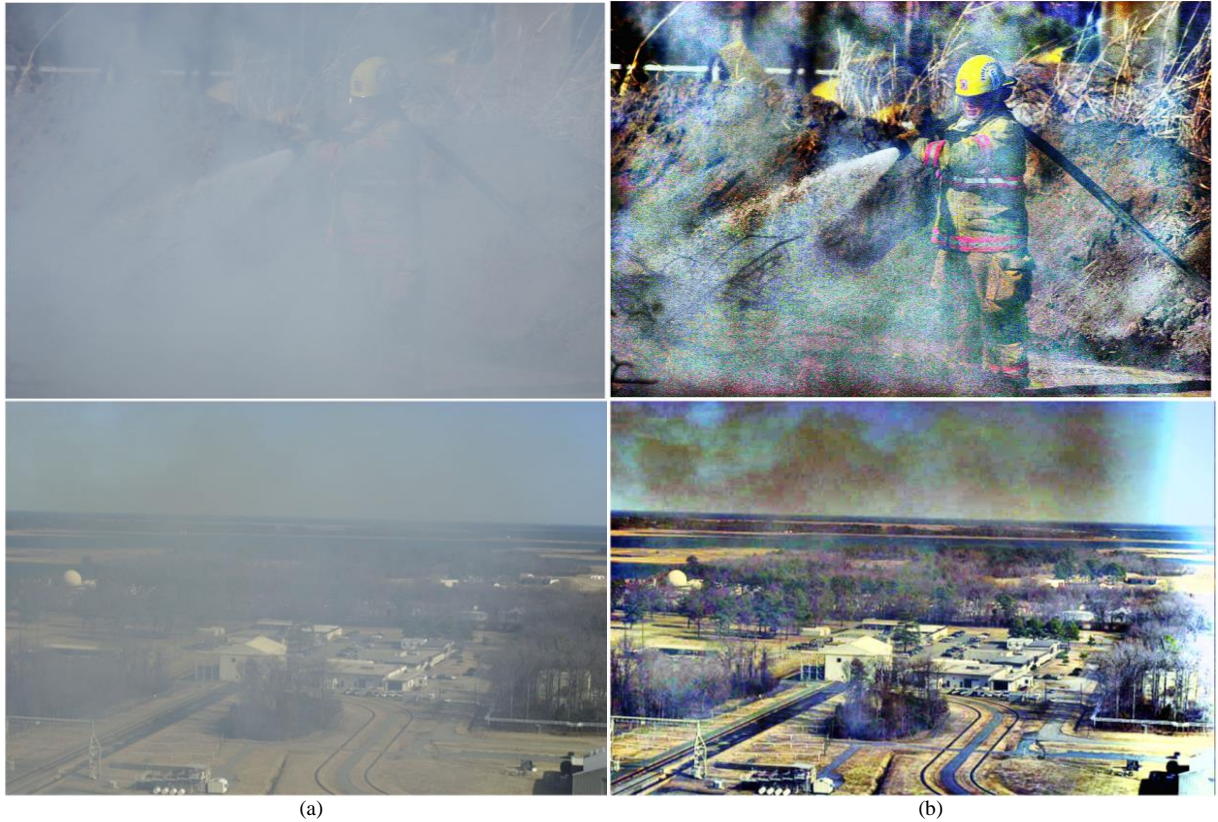

(a)          (b)

**Fig. 17** (a) Original hazy image (b) enhanced image using PA

## Conclusion

This report has presented the results from the work described in **[42]** .